\documentclass[lettersize,journal]{IEEEtran}
\usepackage{amsmath,amsfonts}
\usepackage{algorithm}
\usepackage{algpseudocode}
\usepackage{array}
\usepackage[caption=false,font=normalsize,labelfont=sf,textfont=sf]{subfig}
\usepackage{textcomp}
\usepackage{stfloats}
\usepackage{url}
\usepackage{verbatim}
\usepackage{graphicx}
\usepackage{cite}
\usepackage[outdir=./imgs]{epstopdf}
\usepackage{multirow}
\usepackage{bm}
\usepackage{amsmath,amssymb,amsfonts}
\usepackage{graphicx}
\usepackage{booktabs}
\usepackage{balance}
\usepackage[dvipsnames]{xcolor}
\usepackage[colorlinks,
            linkcolor=blue,
            urlcolor=blue,
            citecolor=blue,
            ]{hyperref}

\newcommand{\etal}{\textit{et al}. }

\newcommand{\eg}{\textit{e}.\textit{g}. }

\begin{document}

\title{Boosting Memory Efficiency in Transfer Learning for High-Resolution Medical Image Classification}

\author{Yijin Huang, Pujin Cheng, Roger Tam, and Xiaoying Tang
        % <-this % stops a space
\thanks{This study was supported by the National Key Research and Development Program of China (2023YFC2415400); the National Natural Science Foundation of China (T2422012, 62071210); the Guangdong Basic and Applied Basic Research (2024B1515020088); the Shenzhen Science and Technology Program (RCYX20210609103056042); the Guangdong Basic and Applied Basic Research (2021A1515220131); the High Level of Special Funds (G030230001, G03034K003).
\textit{(Corresponding authors: Dr. Xiaoying Tang; Dr. Roger Tam)}.}
\thanks{Yijin Huang is with Department of Electronic and Electrical Engineering, Southern University of Science and Technology, Shenzhen 518055, China, and also with School of Biomedical Engineering, The University of British Columbia, Vancouver, BC V6T 1Z3, Canada (e-mail: yijinh@student.ubc.ca).}
\thanks{Pujin Cheng is with Department of Electronic and Electrical Engineering, Southern University of Science and Technology, Shenzhen 518055, China, and also with Department of Electrical and Electronic Engineering, The University of Hong Kong, Hong Kong, China. (e-mail: chengpj@connect.hku.hk).}
\thanks{Roger Tam is with School of Biomedical Engineering, The University of British Columbia, Vancouver, BC V6T 1Z3, Canada (e-mail: roger.tam@ubc.ca).}
\thanks{Xiaoying Tang is with Department of Electronic and Electrical Engineering, Southern University of Science and Technology, Shenzhen 518055, China, and also with Jiaxing Research Institute, Southern University of Science and Technology, Jiaxing 314001, China (e-mail: tangxy@sustech.edu.cn).}
}

\markboth{Preprint, April~2025}%
{Shell \MakeLowercase{\textit{et al.}}: A Sample Article Using IEEEtran.cls for IEEE Journals}


\maketitle

\begin{abstract}
The success of large-scale pre-trained models has established fine-tuning as a standard method for achieving significant improvements in downstream tasks. However, fine-tuning the entire parameter set of a pre-trained model is costly. Parameter-efficient transfer learning (PETL) has recently emerged as a cost-effective alternative for adapting pre-trained models to downstream tasks. Despite its advantages, the increasing model size and input resolution present challenges for PETL, as the training memory consumption is not reduced as effectively as the parameter usage. In this paper, we introduce Fine-grained Prompt Tuning plus (FPT+), a PETL method designed for high-resolution medical image classification, which significantly reduces the training memory consumption  compared to other PETL methods. FPT+ performs transfer learning by training a lightweight side network and accessing pre-trained knowledge from a large pre-trained model (LPM) through fine-grained prompts and fusion modules. Specifically, we freeze the LPM of interest and construct a learnable lightweight side network. The frozen LPM processes high-resolution images to extract fine-grained features, while the side network employs corresponding down-sampled low-resolution images to minimize the memory usage. To enable the side network to leverage pre-trained knowledge, we propose fine-grained prompts and fusion modules, which collaborate to summarize information through the LPM's intermediate activations. We evaluate FPT+ on eight medical image datasets of varying sizes, modalities, and complexities. Experimental results demonstrate that FPT+ outperforms other PETL methods, using only 1.03\% of the learnable parameters and 3.18\% of the memory required for fine-tuning an entire ViT-B model. Our code is available on https://github.com/YijinHuang/FPT.
\end{abstract}

\begin{IEEEkeywords}
Parameter-efficient transfer learning, Memory-efficient transfer learning, Large-scale pre-trained models, High-resolution medical image classification.
\end{IEEEkeywords}

\section{Introduction}
Recently, large pre-trained models (LPMs) have made remarkable achievements across diverse domains \cite{Radford2019, Devlin2019, Yang2019, Radford2021, Zhang2023biomedclip, Kirillov2023, Cheng2023}. Utilizing the technique of transfer learning \cite{Weiss2016}, pre-trained models can be effectively adapted to specific downstream tasks by initializing task-specific models with weights from a pre-trained model of interest, followed by training on task-specific datasets. However, as model sizes grow rapidly (\eg ViT-B, a variant of vision transformer (ViT) \cite{Dosovitskiy2021}, has 86 million parameters), fine-tuning the entire parameter set of an LPM has become very costly, demanding substantial training resources such as GPU memory and significantly limiting its practical applicability.

\begin{figure}[t]
	\centering
	\includegraphics[width=\columnwidth]{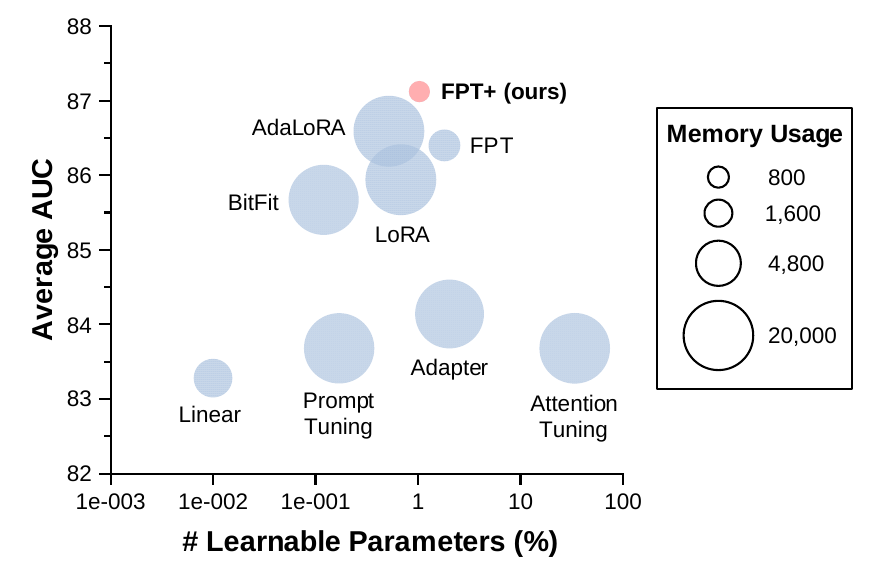}
	\caption{We utilize vision transformer (ViT-B) \cite{Dosovitskiy2021}, pre-trained on ImageNet-21K \cite{Ridnik2021}, as the encoder. The reported performance is the average of performance metrics across eight medical image datasets. The size of the dot denotes the memory consumption during transfer learning. Our proposed method, FPT+, distinguishes itself from other PETL methods by achieving the best balance between performance and efficiency.}
	\label{tradeoff}
\end{figure}

In response, the concept of PETL \cite{Ding2023} has emerged. PETL offers a strategic approach for transferring pre-trained models by selectively updating a small subset of the pre-trained parameters or introducing a modest number of additional parameters specific to the new task while freezing the majority of the pre-trained parameters.

PETL initially gained substantial attention within the natural language processing (NLP) community before finding its application in computer vision (CV) where various PETL methods have shown effectiveness across diverse tasks and datasets. While these approaches have achieved notable success, they were primarily designed for natural images and the field of medical image analysis has not yet fully benefited from such advances \cite{Dutt2023}. The unique challenges of medical image analysis require specialized PETL methods due to the domain gap between natural and medical images \cite{Shen2017, Zu2024, Kim2024}.

\begin{figure}[t]
	\centering
	\includegraphics[width=\columnwidth]{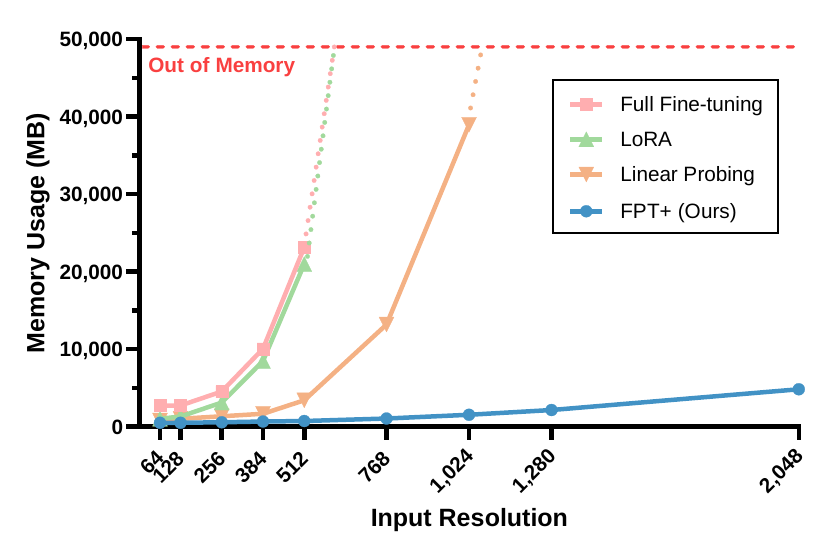}
	\caption{High resolution comes at the cost of heightened GPU memory consumption. The red line represents the maximum memory size of the NVIDIA RTX A6000 GPU used in our experiments. Our proposed FPT+ makes training on super high-resolution images feasible.}
	\label{memory}
\end{figure}

In natural images, the objects of interest typically occupy a significant portion of the image and exhibit clear and noticeable attributes (e.g., shape and color). In contrast, the diagnostic cues within medical images are often tiny and spread throughout the entire visual field, making fine-grained information crucial for accurate medical image analysis \cite{Huang2024}. Providing such fine-grained information often requires high-resolution input images \cite{Huang2023}. However, as illustrated in Fig.~\ref{memory}, this preference for high resolution comes at the cost of increased GPU memory consumption and higher training expenses. For example, a ViT with a patch size of 16 divides a 224 $\times$ 224 image into 196 patches, whereas a 512 $\times$ 512 image generates 1,024 patches. This sharp rise in computational demand makes fine-tuning LPMs both expensive and, in some cases, infeasible on devices with limited GPU memory.

Moreover, it is worth noting that existing LPMs are primarily pre-trained on large natural image datasets like ImageNet \cite{Deng2009}. Attempting to bridge the knowledge gap between natural and medical images by merely updating a small subset of the pre-trained parameters or introducing additional parameters within a frozen LPM may not be sufficient to address the significant differences between these two domains \cite{Zu2024}.

In this work, we propose a novel parameter and memory-efficient PETL method, namely \textbf{F}ine-grained \textbf{P}rompt \textbf{T}uning \textbf{plus} (FPT+). FPT+ aims to enhance the efficiency of PETL specifically for high-resolution medical images. Existing PETL methods typically involve training a subset of parameters within the LPM of interest. However, updating these parameters still requires gradients' computation through the frozen layers during back-propagation, significantly increasing memory consumption. To address this issue, FPT+ employs a lightweight additive network inspired by the concept of a side network \cite{Zhang2020, Sung2022}. This design treats the entire pre-trained model as a frozen feature extractor, eliminating the need for back-propagation through the LPM. To leverage pre-trained knowledge from the LPM, FPT+ introduces the concept of fine-grained prompts and fine-grained fusion modules as the bridging components. Fine-grained prompts consist of a small set of trainable embeddings prepended to the input sequence of the side network. Instead of engaging the entire input sequence for interaction with the fine-grained features from the pre-trained model, the fusion process involves only the fine-grained prompts through cross-attention mechanisms. These prompts subsequently convey fine-grained information to the side network during forward propagation, allowing the integration of pre-trained knowledge into the learnable side network at a low cost.

Additionally, considering that long input sequences from high-resolution images result in substantial intermediate activations, which can be computationally expensive, FPT+ employs important tokens selection to mitigate this issue. Based on the prior knowledge that medical images of the same modality often share similar anatomical characteristics and have small objects of interest (\eg lesions and vessels in fundus images, and lungs in chest X-rays), FPT+ selectively fuses features from the intermediate activations to reduce GPU memory consumption. Preloading techniques are also proposed to improve the training efficiency by pre-storing the LPM's intermediate features.

To validate the effectiveness of the proposed FPT+, we adopt eight publicly accessible medical image datasets with varying sizes, modalities, and complexities. Experimental results demonstrate that FPT+ achieves the best performance-efficiency balance, outperforming other state-of-the-art (SOTA) PETL methods with only 1.03\% of the trainable parameters and 3.18\% of the memory cost required for fine-tuning an entire model.

This paper's main contributions are summarized as follows:
\begin{itemize}
    \item We propose a novel PETL method, namely FPT+, for high-resolution medical image classification. FPT+ maximizes the memory efficiency of transfer learning through asymmetric input and important tokens selection. The effectiveness and efficiency of adapting out-of-domain knowledge are achieved by our fine-grained prompts and fusion modules.
    \item FPT+ sets a new benchmark in memory efficiency for PETL methods, achieving competitive performance utilizing only 3.18\% of the memory cost required for fine-tuning an entire model. As shown in Fig.~\ref{memory}, FPT+ significantly extends the applicability of transfer learning within the realm of high-resolution inputs.
    \item Extensive experiments are conducted on eight medical image datasets with various sizes, modalities and complexities, demonstrating the effectiveness of FPT+ by achieving the best balance between performance and parameter/memory efficiency. Additionally, resource consumption is comprehensively evaluated to establish the superior training efficiency of FPT+.
    \item We evaluate different LPMs for transfer learning, including ImageNet-based supervised and self-supervised LPMs, as well as a medical-specific pre-trained LPM, emphasizing the generalizability of FPT+ across different LPM scales and different pre-trained weights.
\end{itemize}

This paper extends our previous work, FPT \cite{Huang2024FPT}. We systematically investigate the impact of several key components in FPT and refine the entire framework to make it simpler and more efficient. Specifically, our extension is presented in several aspects: (1) We reduce the number of layers in the side network and its corresponding fusion modules to significantly decrease memory consumption; (2) We redesign the fusion module by replacing in-projector with layer normalization and align the initial hidden dimension of the fine-grained prompts with the frozen LPM; (3) All fusion modules share the same weights for the out-projector, further reducing the number of learnable parameters; (4) The new design improves the effectiveness of learning pre-trained knowledge, allowing smaller low-resolution inputs for the side network, thus further reducing memory consumption; (5) To verify the generalizability of FPT+, we validate the framework using four additional datasets; (6) We evaluate the scalability and feasibility of FPT+ across different input resolutions, pre-trained models of different scales, configurations, and weights; (7) Additional efficiency metrics, including training time, throughput, and power usage, are evaluated to comprehensively assess FPT+'s resource consumption and training efficiency; (8) Extensive ablation studies are conducted to qualitatively and quantitatively analyze the impact of each component in FPT+ on performance and efficiency.

\section{Related Work}
\subsection{Parameter-efficient Transfer Learning}
In recent years, the practice of fine-tuning expansive pre-trained models has shown encouraging potential across a range of tasks in the field of NLP. With the rapidly increasing sizes of pre-trained models, typically transformers \cite{Vaswani2017, brown2020language}, the concept of PETL has emerged as a strategic solution to mitigate the computational burden in adapting large-scale pre-trained models to downstream tasks.

LoRA-based methods \cite{Hu2022, Zhang2023} freeze pre-trained weights and inject trainable rank decomposition matrices into each layer of a transformer to reduce the number of trainable parameters. Prompt tuning \cite{Lester2021} appends the input with a trainable tensor, termed prompt, while freezing all other parameters. Adapter \cite{Houlsby2019} introduces the idea of adding new learnable modules among layers of a frozen pre-trained model to adapt to new downstream tasks. BitFit \cite{Elad2022} tunes only the bias term of a pre-trained model to effectively adapt the model to downstream tasks.

Drawing inspiration from NLP advancements, large-scale pre-trained models for CV are also receiving growing attention. ViTs are widely adopted as the backbone due to their modeling power and scalability to model and data size \cite{Zhang2023, zhou2023foundation, Ma2024, 10580962}. While most PETL methods were initially designed for transformers, recent works have demonstrated their effectiveness in ViTs as well.

There are also several PETL methods specifically designed for ViT \cite{Chen2022, Jia2022, Zhou2022, Liu2023, Gong2023}. Attention tuning \cite{Touvron2022} fine-tunes only the attention layers to adapt ViTs to various downstream tasks. Vision prompt tuning \cite{Jia2022} expands upon the concept of prompt tuning by introducing learnable prompts into each block of a frozen pre-trained ViT. Side tuning \cite{Zhang2020} employs an additive side network, which combines features from a frozen pre-trained model using a simple summation operation facilitated by a trainable gate. Liu \etal \cite{Liu2023} propose Explicit Visual Prompting to improve fine-tuning on segmentation tasks by taking the high-frequency components in the input as the prompts. CoOp \cite{Zhou2022} and CoCoOp \cite{Zhou2022cocoop} are proposed to specifically adapt CLIP-like vision-language pre-trained models to downstream image recognition tasks. Zhao \etal \cite{Zhao2024} leverage task-specific information to select salient channels and then fine-tune only those channels to enhance parameter efficiency.

Unlike other PETL methods, our proposed FPT+ aim to improve both parameter and memory efficiency during transfer learning, which is a specific and common requirement in the field of medical image analysis.

\subsection{PETL for Medical Image Analysis}
While several PETL methods have emerged in the realm of CV, the application of PETL to ViTs for medical images remains relatively limited, despite the efficiency requirement inherent in the medical domain \cite{Dutt2023}. Raman \etal \cite{Dutt2023} comprehensively evaluate the performance of different PETL methods for medical image analysis. Rodríguez \etal \cite{Julio2023} study adapters designed for dense prediction tasks to improve medical image segmentation. Dynamic Visual Prompt Tuning \cite{He2023DVPT} is designed to learn sample-specific and domain-specific features to enhance the fine-tuning performance in the medical domain. Zu \etal \cite{Zu2024} propose Embedded Prompt Tuning by embedding prompt tokens into expanded channels for few-shot medical image classification tasks. Wu \etal \cite{Wu2023} combine the SAM model \cite{Kirillov2023} with adapters, which outperforms existing SOTA methods on several downstream medical image segmentation tasks.

Different from our previous work, FPT \cite{Huang2024FPT}, we simplify and enhance its framework for greater efficiency. Key improvements include reducing memory consumption through optimized side networks, fusion modules and weights sharing strategy, and enabling effective learning with smaller inputs. Unlike other PETL methods for medical image analysis, our proposed FPT+ framework aims to simultaneously address the challenge of unaffordable parameter as well as memory utilization when transferring LPMs to high-resolution medical image classification tasks.

\begin{figure*}[t]
	\centering
	\includegraphics[width=\textwidth]{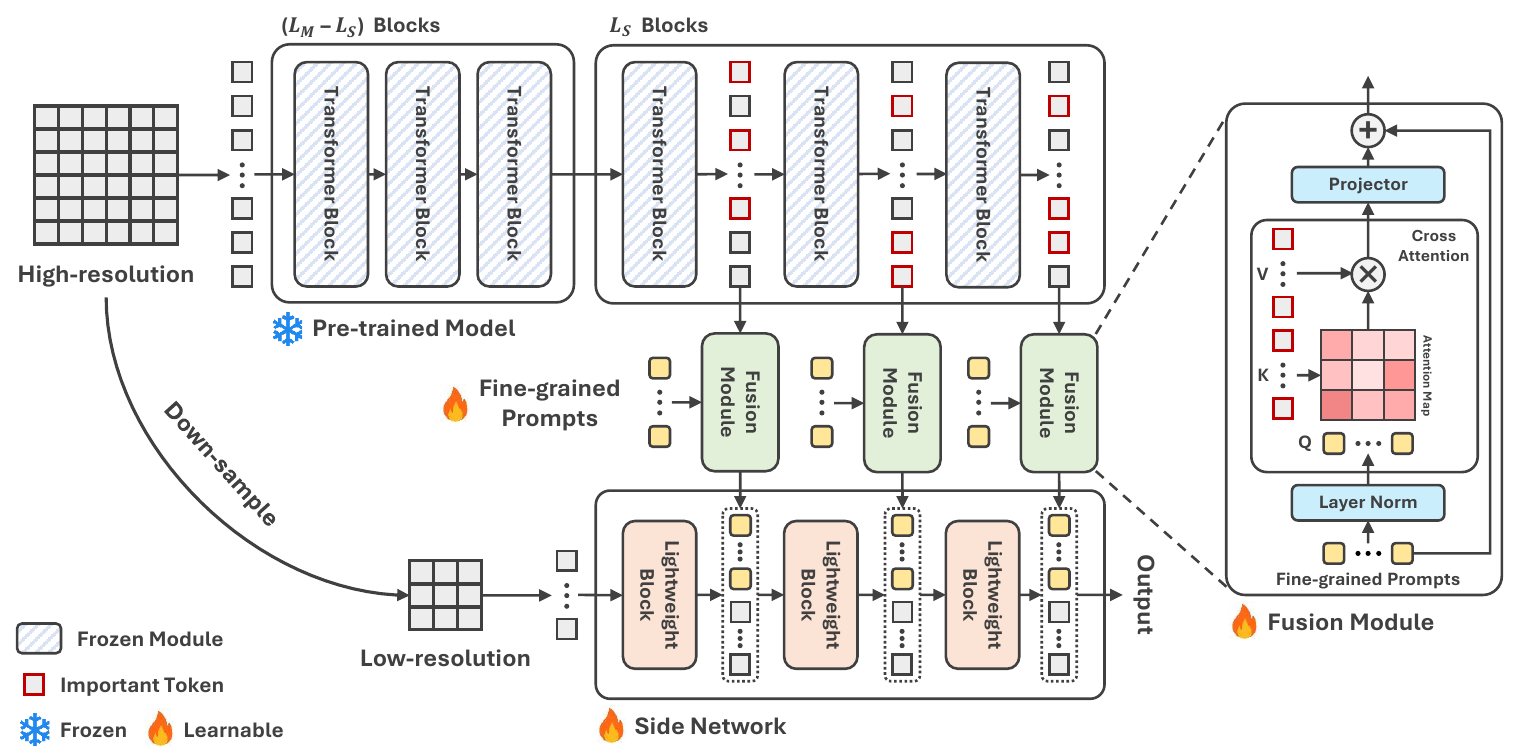}
	\caption{Our FPT+ framework comprises the following components: (a) The large-scale pre-trained model remains frozen and receives high-resolution images as the input. A lightweight, learnable side network is employed to integrate knowledge from the pre-trained model into new downstream tasks through the utilization of fine-grained prompts and fusion modules. (b) The fine-grained fusion module operates by taking fine-grained prompts and important features extracted from the pre-trained model as inputs. Prompts and features are fused through feature alignment and cross-attention.}
	\label{framework}
\end{figure*}

\section{Method}
In this work, we propose FPT+, a parameter and memory efficient transfer learning method for high-resolution medical image classification. FPT+ can: 1) efficiently integrate fine-grained information from a pre-trained model into a new downstream model; and 2) effectively adapt knowledge from an out-of-domain pre-trained model. As shown in Fig.~\ref{framework}, a separate and lightweight side network is first employed to learn and infer at an affordable training cost in terms of both the number of trainable parameters and memory consumption. Then, fine-grained prompts and fusion modules are proposed to convey fine-grained pre-trained knowledge to the side network effectively and efficiently. Asymmetric input and important tokens selection are introduced to further reduce the computational burden. The motivation, benefits, and functioning of each component are delineated in subsequent sections.

\subsection{Vision Transformer}
In this section, we provide a brief overview of ViT \cite{Dosovitskiy2021}. A conventional ViT architecture consists of a linear projection layer followed by multiple transformer blocks. For a given image $x \in \mathbb{R}^{H \times W \times C}$ with spatial dimensions $(H, W)$ and channel dimension $C$, the initial step involves partitioning the image into non-overlapping 2D patches of size $P \times P$. Here, $P$ represents the patch size specific to the ViT of interest. Subsequently, these 2D patches are flattened and transformed into a latent embedding space with hidden dimension $D$, facilitated by a linear projection layer. This process yields an input sequence $\bm{z}_0 = {z_p^i \in \mathbb{R}^D | i=1,...,N}$ with a length of $N = HW / P^2$. Each element in this input sequence is termed a token. The input is then processed by multiple transformer blocks, each consisting of a multi-headed self-attention (MSA) layer and a multi-layer perceptron (MLP) block:
\begin{align}
    &\bm{z}_l^\prime = \text{MSA}(\bm{z}_{l-1}) + \bm{z}_{l-1}, \\
    &\bm{z}_l = \text{MLP}(\bm{z}_l^\prime) + \bm{z}_l^\prime,
\end{align}
where $l$ indexes layers of the ViT. The class token and layer normalization operations are omitted in the above formulas for simplification.

Within each self-attention layer, individual tokens in the input sequence are mapped to Q (query), K (key), and V (value) representations for computations in the attention module (Attn) \cite{Vaswani2017} as follows:
\begin{align}
    \text{MSA}(\bm{z}) &= \text{Attn}(Q, K, V) \nonumber \\
    &= \text{Attn}(f_Q(\bm{z}), f_K(\bm{z}), f_V(\bm{z})),
\end{align}
where $f_Q$, $f_K$, and $f_V$ denote the mapping functions in the attention module.

\subsection{Side Tuning}
Unlike popular PETL methods which introduce supplementary modules into transformer blocks, our approach, drawing inspiration from \cite{Zhang2020, Sung2022}, adopts a lightweight and separate side network. The side network takes advantages of the LPM by utilizing intermediate features from the frozen pre-trained model. As depicted in part (a) of Fig.~\ref{framework}, the lightweight ViT on the right is a scaled-down variant of the LPM on the left in terms of hidden dimensions and the number of layers, serving as the side network. The hidden dimensions of the side network are $1/k$ times that of the original, and its parameter weights are initialized randomly. Here, $k$ represents a reduction factor. The side network is shallower than the LPM in terms of the number of layers.

To utilize knowledge from the LPM, a novel fusion module $\mathcal{F}$ is introduced to fuse the intermediate features of the LPM and those of the side network at each layer, which will be elaborated upon in the subsequent section. Specifically, given the LPM $M$ with $L_M$ layers and the side network $S$ with $L_S$ layers, respectively parameterized by $\theta_M$ and $\theta_S$, the intermediate activation $\bm{z}_S^l$ of layer $l$ in the side network is obtained as follows:
\begin{align}
    &\bm{z}_M^{l^\prime} = \theta_M^{l^\prime}(\bm{z}_M^{l^\prime - 1}), \\
    &\bm{z}_S^l = \theta_S^l(\mathcal{F}(\bm{z}_S^{l-1}, \bm{z}_M^{l^\prime})) \label{zs},
\end{align}
where $l^\prime = L_M - L_S + l$ and $l \in [1, L_S]$, and $\bm{z}_S^{L_S}$ is considered as the final output of the entire framework.

\begin{figure}[t]
	\centering
	\includegraphics[width=0.75\columnwidth]{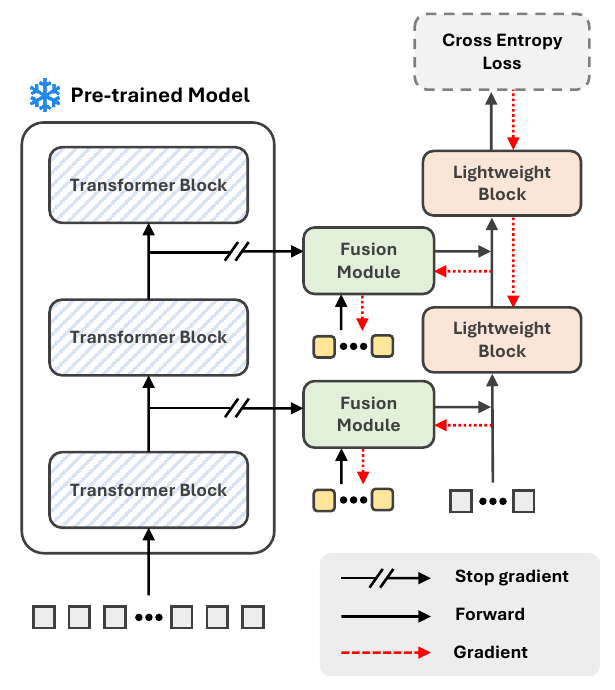}
	\caption{There are no learnable parameters inside the LPM, eliminating the need for gradient computation inside the heavy LPM.}
	\label{gradient}
\end{figure}

Adoption of the side network not only reduces the number of trainable parameters through its lightweight design but also mitigates memory expenses during the training phase. As shown in Fig.~\ref{gradient}, given the separation between the side network and the pre-trained model, the forward pass of the frozen pre-trained model does not involve any learnable parameter. Consequently, updating the side network prevents the need for memory-intensive back-propagation inside the heavy LPM, resulting in notable memory savings during training.

\subsection{Asymmetric Input}
Fine-grained information holds significant importance in medical image analysis. To extract such detailed information, one straightforward approach involves increasing the resolution of the input image. However, due to the mechanics of input patchification and the self-attention layer in the ViT architecture, increasing the input resolution leads to a dramatic rise in memory consumption and training cost. For instance, in a ViT with a patch size of 16, increasing the input resolution from 224 to 512 results in an expansion of the input sequence length from 196 to 1,024. Furthermore, the dimensions of the self-attention maps in the transformer block escalate from $196^2$ to $1,024^2$. This considerable training overhead presents challenges for fine-tuning on high-resolution datasets. 

Although high-resolution inputs make the training cost unfeasible, employing a pre-trained model for inference with high-resolution images remains practical. We propose here an asymmetric input strategy within the FPT+ framework. Specifically, the input images $I_h$ for the frozen pre-trained model are of high resolution, whereas the images fed to the learnable side network are low-resolution images $I_l$ that get down-sampled from the high-resolution ones. Therefore, the Equation~\ref{zs} becomes
\begin{align}
    &\bm{z}_{S,I_l}^l = \theta_S^l(\mathcal{F}(\bm{z}_{S,I_l}^{l-1}, \bm{z}_{M,I_h}^{l^\prime})).
\end{align}

\subsection{Fine-grained Prompts and Fusion Modules}
In this section, we introduce our proposed fusion modules $\mathcal{F}$ and fine-grained prompts. To fuse features from the LPM and those from the side network, a simple approach is to use an MLP layer to process and align the hidden dimensions of those features, and then concatenate them into the intermediate sequence of the side network. However, this approach will dramatically increase the length of the token sequence of the side network, significantly increasing memory consumption. Furthermore, popular LMPs are generally pre-trained on natural images, and since the pre-trained model remains frozen during training, integrating this out-of-domain knowledge might not seamlessly adapt to new medical downstream tasks. Moreover, directly fusing features across diverse scales and domains poses challenges for the side network's ability to effectively learn from its input images.

To tackle these challenges, as illustrated in part~(b) of Fig.~\ref{framework}, we introduce a small set of learnable embeddings $\bm{p}$ named fine-grained prompts and utilize a cross-attention mechanism \cite{Chen2021, Vaswani2017} to summarize pre-trained knowledge from the LPM. Unlike prompt tuning \cite{Lester2021, Jia2022} which directly uses prompts as part of the input sequence, our proposed fine-grained prompts serve as a bridge linking the frozen LPM and the side network. Specifically, in the context of cross-attention, we reuse the key $K_M$ and the value $V_M$ from the self-attention layer of the LPM, while the fine-grained prompts serve as the query. The cross-attention module allows the prompts to retrieve tokens of interest and summarize fine-grained information.

Subsequently, these fine-grained prompts are concatenated with the intermediate sequence of the side network, interacting with other tokens to share the fused pre-trained knowledge. The processing of the fusion module goes as follows:
\begin{align}
    \mathcal{F}(\bm{z}_S, \bm{z}_M) &= [\bm{z}_S, f_{\text{out}}(\text{CA}(\bm{p}, \bm{z}_M)) + \bm{p}], \\
    \text{CA}(\bm{p}, \bm{z}_M) &= \text{Attn}(Q, K, V) \nonumber \\
                         &= \text{Attn}(\text{LN}(\bm{p}), K_M, V_M),
\end{align}
where notations for the input source and the layer index are omitted for simplification. The $\text{CA}(\cdot)$ and $\text{LN}(\cdot)$ denote the cross-attention module and the layer normalization operation, respectively. The $[\cdot, \cdot]$ denotes the concatenation operation, and the out-projector $f_{\text{out}}$ is a linear layer to align the hidden dimension of the fine-grained prompts to that of the side network. The out-projector is shared across all fusion modules to further increase efficiency. Notably, the number of the fine-grained prompts is considerably smaller than that of the input sequence, leading to reduced training cost. They are removed after each layer's forward processing in the side network to maintain consistency of the sequence length across layers. The pseudo-code in Algorithm~\ref{alg:fusion_code} for the fusion module is provided to further clarify its implementation details.

\begin{algorithm}[t]
\caption{Fusion Module with Fine-grained Prompts}
\label{alg:fusion_code}
\begin{algorithmic}[1]
\State \textbf{Input:} Side network tokens $z_S$, fine-grained prompts $p$, LPM-projected keys $K_M$ and values $V_M$
\State \textbf{Output:} Fused token sequence $\tilde{z}_S$
\vspace{2mm}

\State $p_{\text{ln}} \gets \mathrm{LN}(p)$ \Comment{Layer normalization}
\State $p_{\text{attn}} \gets \mathrm{Attn}(p_{\text{ln}}, K_M, V_M)$ \Comment{Cross-attention(Q, K, V)}
\State $p_{\text{out}} \gets f_{\text{out}}(p_{\text{attn}})$ \Comment{Linear projection of attention output}
\State $\tilde{p} \gets p_{\text{out}} + p$ \Comment{Residual connection}
\State $\tilde{z}_S \gets [z_S, \tilde{p}]$  \Comment{Concat prompts to side network tokens}

\vspace{2mm}
\State \Return $\tilde{z}_S$
\end{algorithmic}
\end{algorithm}

\subsection{Important Tokens Selection}
Despite no need for back-propagation from the frozen LPM and the great memory savings provided by the proposed fusion modules, integrating lengthy intermediate features produced by high-resolution inputs from the LPM still introduces a significant memory load. Inspired by the prior knowledge that medical images of the same modality often exhibit similar anatomical characteristics and that the diagnostic clues generally occupy only a small proportion of the entire image, we identify tokens containing diagnostic information within the input sequence of the pre-trained model as the main contributors for downstream tasks.

In such context, we present a straightforward yet effective approach to mitigate the training expenses: we select important tokens, and only features associated with those important tokens are passed to the fusion modules. Specifically, we assess the tokens' importance based on the self-attention maps \cite{Dosovitskiy2021} in the LPM. The importance score of a token is defined as the average attention score of other tokens upon it. Subsequently, we select the top $m\%$ tokens with the highest scores as the important tokens, which are delivered to the side network through the fusion modules. This strategy alleviates the overhead introduced by the lengthy tokens from the high-resolution inputs, while preserving the discriminative features essential for downstream tasks.

\subsection{Fine-grained Features Preloading}
To further enhance the training efficiency, we choose not to employ any data augmentation for the high-resolution input of the frozen LPM, ensuring that features associated with an image of interest remain constant throughout training. Consequently, it becomes feasible to pre-store these features prior to training, resulting in notable reductions in the training cost. Leveraging the benefits of important tokens selection, the demand for storage and the overhead linked to data loading are also considerably reduced, thereby collectively accelerating the training process. Although the high-resolution input of the LPM remains fixed, we apply data augmentation to the low-resolution input of the side network to maintain the diversity of the training samples. Note that the preloading technique is exclusive to FPT+, because the asymmetric input design allows for the application of preloading on the high-resolution input while avoiding overfitting and performance degradation through data augmentation on the low-resolution input.

\section{Experiments}
\subsection{Datasets}
We evaluate FPT+ using eight medical image datasets for classification tasks across four modalities, including fundus images (DDR \cite{Li2019} and Messidor-2 \cite{Messidor2}), dermoscopic images (ISIC 2016 \cite{Gutman2016} and ISIC 2018 \cite{Codella2019}), mammography images (DDSM \cite{Lee2017} and CMMD \cite{Cui2021}), and chest X-ray images (COVID \cite{Siddhartha2021} and CHNCXR \cite{Jaeger2014}). The dataset sizes range from 662 to 12,522 samples, with classification categories varying from 2 to 7. When official splits for training, validation, and testing are available, they are adopted. Otherwise, a random partition of 70\%/10\%/20\% is employed for training, validation, and testing, respectively.

\subsection{Training and Evaluation Setup}
\subsubsection{Experiment Setup}
\label{setup}
In this study, we employ two popular variants of ViT \cite{Dosovitskiy2021}, specifically ViT-B(ase) and ViT-L(arge), as our large-scale pre-trained models, with 86 million and 304 million parameters, respectively. To evaluate the generalizability of different PETL methods, we experiment with three types of pre-trained weights: one obtained through supervised learning on ImageNet-21K \cite{Ridnik2021}, another via self-supervised learning using DINO \cite{Caron2021} pre-trained on ImageNet-21K, and the third using BioMedCLIP \cite{Zhang2023biomedclip}, a vision-language model pre-trained on various medical imaging modalities. These models are widely recognized for their robust generalizability in both natural and medical image tasks. The default patch size used in this work for ViTs is 16, with additional evaluations conducted on ViT-B using a patch size of 8. All methods are trained for 20 epochs with a mini-batch size of 16 on an NVIDIA RTX A6000 GPU (48GB) using PyTorch of version 2.0.0 \cite{paszke2017automatic}. We adopt a cosine annealing schedule for the learning rate during training. ViTs are trained using the AdamW optimizer with cross-entropy as the loss function for all datasets. Each experiment is repeated 5 times, and we report the mean $\pm$ standard deviation values.

If not specified, all compared methods run at a resolution of $512 \times 512$ with ImageNet-based supervised LPM. In the context of FPT+, the high resolution for the LPM remains $512 \times 512$, while the low resolution for the side network is set to $128 \times 128$. The number of fine-grained prompts is set to 16, and the reduction factor $k$ is set to 8. For important tokens selection, the top 20\% of the important tokens are retained for utilization within the fusion modules. The number of layers in the side network $L_S$ is set to 6, which is half the number of layers in the LPM.

For compared methods, LoRA \cite{Hu2022} and AdaLoRA \cite{Zhang2023} use a scale factor ($\alpha$) of 16, with LoRA layers applied to the query and value weights in the ViT. Prompt tuning is reproduced following the VPT-deep design \cite{Jia2022}, with 16 prompts. To ensure fair comparisons across methods, grid searching is utilized to determine the optimal set of hyperparameters, including the learning rate and the weight decay, based on performance on the validation set. All methods and the side network of FPT+ are trained using the same data augmentations, including cropping, horizontal flipping, and color distortion.

\subsubsection{Evaluation Metrics}
We use the Area Under the Receiver Operating Characteristic Curve (AUC) to evaluate the classification performance on each dataset. To evaluate the performance and efficiency balance, we adopt the metrics proposed in FPT\cite{Huang2024FPT}, which extend from the performance-efficiency metric \cite{Li2022, He2023}. Specifically, two metrics, namely performance-parameter-efficiency (PPE) and performance-memory-efficiency (PME), are utilized to assess the impact of the number of learnable parameters and GPU memory consumption on efficiency during transfer learning, respectively. They are defined as
\begin{align}
	\text{PPE} &= \text{score} \times e^{-\log_{10}(r + 1)}, \\
    \text{PME} &= \text{score} \times e^{-\log_{10}(m + 1)},
\end{align}
where $\text{score}$ is the average performance across all datasets, $r$ is the ratio of learnable parameters to all parameters, and $m$ is the ratio of GPU memory requirement of a method of interest to that of fine-tuning the entire LPM.

\begin{table*}[t]
    \centering
    \renewcommand{\arraystretch}{1.2}
    \caption{Comparison results with state-of-the-art PETL methods across all evaluation datasets. 'Params' denotes the ratio of learnable parameters to the total number of parameters. 'Mem.' denotes the memory usage (MB). 'Avg.', 'PPE', and 'PME' denote the average performance across all datasets, performance-parameter-efficiency, and performance-memory-efficiency metrics. The best and second-best results are indicated in bold and underlined formatting, respectively.}
    \resizebox{\textwidth}{!}{%
        \begin{tabular}{lcccccccccccccccccc}
        \toprule
                                                & \multicolumn{2}{c}{Computing cost}                         &                         & \multicolumn{2}{c}{Fundus image}                                          & \multicolumn{1}{c}{}    & \multicolumn{2}{c}{Dermoscopic image}                                     & \multicolumn{1}{c}{}    & \multicolumn{2}{c}{Mammography}                                           & \multicolumn{1}{c}{}    & \multicolumn{2}{c}{Chest X-ray}                                           &                         & \multicolumn{1}{c}{}                       & \multicolumn{1}{c}{}                      & \multicolumn{1}{c}{}                      \\ \cline{2-3} \cline{5-6} \cline{8-9} \cline{11-12} \cline{14-15}
        \multirow{-2}{*}{Method}                & Params. $\downarrow$                     & Mem. $\downarrow$                         &                         & \multicolumn{1}{c}{DDR}             & \multicolumn{1}{c}{Messidor-2}       & \multicolumn{1}{c}{}    & \multicolumn{1}{c}{ISIC2018}        & \multicolumn{1}{c}{ISIC2016}        & \multicolumn{1}{c}{}    & \multicolumn{1}{c}{DDSM}            & \multicolumn{1}{c}{CMMD}            & \multicolumn{1}{c}{}    & \multicolumn{1}{c}{COVID}           & \multicolumn{1}{c}{CHNCXR}              &                         & \multicolumn{1}{c}{\multirow{-2}{*}{Avg. $\uparrow$}} & \multicolumn{1}{c}{\multirow{-2}{*}{PPE $\uparrow$}} & \multicolumn{1}{c}{\multirow{-2}{*}{PME $\uparrow$}} \\ \hline
        \multicolumn{19}{c}{\textit{Supervised (ImageNet)}}                                                                                                                                                                                                                                                                                                                                                                                                                                                                                                                                                                                                                                         \\ \hline
        {\color[HTML]{9B9B9B} Full fine-tuning} & {\color[HTML]{9B9B9B} 100}  & {\color[HTML]{9B9B9B} 23,128} & {\color[HTML]{9B9B9B} } & {\color[HTML]{9B9B9B} 89.79 $\pm$ 0.76} & {\color[HTML]{9B9B9B} 86.87 $\pm$ 0.53} & {\color[HTML]{9B9B9B} } & {\color[HTML]{9B9B9B} 96.65 $\pm$ 0.29} & {\color[HTML]{9B9B9B} 83.22 $\pm$ 1.02} & {\color[HTML]{9B9B9B} } & {\color[HTML]{9B9B9B} 92.49 $\pm$ 0.34} & {\color[HTML]{9B9B9B} 66.26 $\pm$ 0.69} & {\color[HTML]{9B9B9B} } & {\color[HTML]{9B9B9B} 99.85 $\pm$ 0.05} & {\color[HTML]{9B9B9B} 95.40 $\pm$ 1.05} & {\color[HTML]{9B9B9B} } & {\color[HTML]{9B9B9B} 88.82}               & {\color[HTML]{9B9B9B} 65.73}              & {\color[HTML]{9B9B9B} 65.73}              \\
        {\color[HTML]{9B9B9B} Linear probing}   & {\color[HTML]{9B9B9B} 0.01} & {\color[HTML]{9B9B9B} 3,416}  & {\color[HTML]{9B9B9B} } & {\color[HTML]{9B9B9B} 80.17 $\pm$ 0.85} & {\color[HTML]{9B9B9B} 79.73 $\pm$ 0.84} & {\color[HTML]{9B9B9B} } & {\color[HTML]{9B9B9B} 93.37 $\pm$ 0.31} & {\color[HTML]{9B9B9B} 81.27 $\pm$ 0.70} & {\color[HTML]{9B9B9B} } & {\color[HTML]{9B9B9B} 80.89 $\pm$ 0.39} & {\color[HTML]{9B9B9B} 60.23 $\pm$ 0.62} & {\color[HTML]{9B9B9B} } & {\color[HTML]{9B9B9B} 99.21 $\pm$ 0.07} & {\color[HTML]{9B9B9B} 91.34 $\pm$ 0.20} & {\color[HTML]{9B9B9B} } & {\color[HTML]{9B9B9B} 83.28}               & {\color[HTML]{9B9B9B} 83.28}              & {\color[HTML]{9B9B9B} 78.44}              \\
        Prompt-tuning \cite{Jia2022}                            & \underline{0.17}                        & 20,582                        &                         & 81.07 $\pm$ 0.59                        & 80.02 $\pm$ 2.34                        &                         & 94.20 $\pm$ 0.20                        & 81.66 $\pm$ 1.03                        &                         & 82.67 $\pm$ 0.34                        & 59.62 $\pm$ 0.83                        &                         & 99.27 $\pm$ 0.03                        & 90.95 $\pm$ 0.82                        &                         & 83.68                                      & 83.62                                     & 63.47                                     \\
        Attention-tuning \cite{Touvron2022}                       & 33.8                        & 20,792                        &                         & 81.99 $\pm$ 1.85                        & 81.87 $\pm$ 1.42                        &                         & 94.40 $\pm$ 0.40                        & 80.11 $\pm$ 1.92                        &                         & 80.67 $\pm$ 2.33                        & 60.03 $\pm$ 1.03                        &                         & 99.58 $\pm$ 0.19                        & \underline{92.07 $\pm$ 2.48}                  &                         & 83.68                                      & 73.74                                     & 63.34                                     \\
        Adapter \cite{Chen2022}                                 & 2.03                        & 19,360                        &                         & \textbf{85.84 $\pm$ 2.14}               & 80.77 $\pm$ 7.48                        &                         & \textbf{95.96 $\pm$ 0.13}               & 79.22 $\pm$ 2.26                        &                         & 80.76 $\pm$ 3.38                        & 61.39 $\pm$ 1.40                        &                         & 99.17 $\pm$ 0.48                        & 90.02 $\pm$ 0.91                        &                         & 84.14                                      & 83.41                                     & 64.61                                     \\
        BitFit \cite{Elad2022}                                  & \textbf{0.12}                        & 20,382                        &                         & 84.31 $\pm$ 0.50                        & 83.81 $\pm$ 1.11                        &                         & 94.84 $\pm$ 0.15                        & 82.35 $\pm$ 0.62                        &                         & 84.77 $\pm$ 0.54                        & 63.84 $\pm$ 1.80                        &                         & \textbf{99.81 $\pm$ 0.03}               & 91.62 $\pm$ 0.95                        &                         & 85.67                                      & 85.63                                     & 65.11                                     \\
        LoRA \cite{Hu2022}                                   & 0.68                        & 20,970                        &                         & \textbf{85.84 $\pm$ 2.42}               & \underline{86.08 $\pm$ 0.95}                  &                         & 95.02 $\pm$ 0.22                        & \underline{84.22 $\pm$ 1.98}                  &                         & 82.26 $\pm$ 4.04                        & 62.26 $\pm$ 1.66                        &                         & 99.69 $\pm$ 0.05                        & 92.13 $\pm$ 1.12                        &                         & 85.94                                      & 85.69                                     & 64.93                                     \\
        AdaLoRA \cite{Zhang2023}                                & 0.52                        & 20,960                        &                         & 85.21 $\pm$ 0.34                        & 86.07 $\pm$ 0.74                        &                         & \underline{95.85 $\pm$ 0.24}                  & \textbf{84.23 $\pm$ 0.65}               &                         & 86.91 $\pm$ 0.30                        & 62.45 $\pm$ 1.46                        &                         & \underline{99.75 $\pm$ 0.04}                  & \textbf{92.25 $\pm$ 0.70}               &                         & \underline{86.59}                                & \underline{86.40}                               & 65.43                                     \\
        FPT \cite{Huang2024FPT}                                    & 1.81                        & \underline{1,824}                         &                         & 84.13 $\pm$ 1.06                        & 84.95 $\pm$ 2.01                        &                         & 93.88 $\pm$ 0.60                        & 80.68 $\pm$ 1.13                        &                         & \underline{90.52 $\pm$ 0.59}                  & \underline{65.35 $\pm$ 0.94}                  &                         & 99.70 $\pm$ 0.30                        & 91.97 $\pm$ 1.36                        &                         & 86.40                                      & 85.73                                     & \underline{83.60}                               \\
        FPT+ (Ours)                                   & 1.03                        & \textbf{736}                          &                         & \underline{85.53 $\pm$ 1.58}                  & \textbf{86.31 $\pm$ 0.66}               &                         & 94.27 $\pm$ 0.32                        & 82.91 $\pm$ 1.25                        &                         & \textbf{90.64 $\pm$ 0.71}               & \textbf{66.33 $\pm$ 0.43}               &                         & 99.71 $\pm$ 0.18                        & 91.30 $\pm$ 0.94                        &                         & \textbf{87.12}                             & \textbf{86.73}                            & \textbf{85.94}                            \\ \hline
        \multicolumn{19}{c}{\textit{DINO (ImageNet)}} \\ \hline
        {\color[HTML]{9B9B9B} Full fine-tuning} & {\color[HTML]{9B9B9B} 100}  & {\color[HTML]{9B9B9B} 23,128} & {\color[HTML]{9B9B9B} } & {\color[HTML]{9B9B9B} 87.33 $\pm$ 0.77} & {\color[HTML]{9B9B9B} 84.06 $\pm$ 1.29} & {\color[HTML]{9B9B9B} } & {\color[HTML]{9B9B9B} 96.44 $\pm$ 0.26} & {\color[HTML]{9B9B9B} 84.46 $\pm$ 1.55} & {\color[HTML]{9B9B9B} } & {\color[HTML]{9B9B9B} 91.19 $\pm$ 0.32} & {\color[HTML]{9B9B9B} 64.70 $\pm$ 0.32} & {\color[HTML]{9B9B9B} } & {\color[HTML]{9B9B9B} 99.92 $\pm$ 0.03} & {\color[HTML]{9B9B9B} 95.31 $\pm$ 1.19} & {\color[HTML]{9B9B9B} } & {\color[HTML]{9B9B9B} 87.93}               & {\color[HTML]{9B9B9B} 65.07}              & {\color[HTML]{9B9B9B} 65.07}              \\
        {\color[HTML]{9B9B9B} Linear probing}   & {\color[HTML]{9B9B9B} 0.01} & {\color[HTML]{9B9B9B} 3,416}  & {\color[HTML]{9B9B9B} } & {\color[HTML]{9B9B9B} 80.77 $\pm$ 0.67} & {\color[HTML]{9B9B9B} 81.71 $\pm$ 0.69} & {\color[HTML]{9B9B9B} } & {\color[HTML]{9B9B9B} 93.44 $\pm$ 0.16} & {\color[HTML]{9B9B9B} 83.96 $\pm$ 0.93} & {\color[HTML]{9B9B9B} } & {\color[HTML]{9B9B9B} 81.06 $\pm$ 0.46} & {\color[HTML]{9B9B9B} 63.64 $\pm$ 1.08} & {\color[HTML]{9B9B9B} } & {\color[HTML]{9B9B9B} 99.40 $\pm$ 0.01} & {\color[HTML]{9B9B9B} 91.84 $\pm$ 0.18} & {\color[HTML]{9B9B9B} } & {\color[HTML]{9B9B9B} 84.48}               & {\color[HTML]{9B9B9B} 84.48}              & {\color[HTML]{9B9B9B} 79.57}              \\
        Prompt-tuning \cite{Jia2022}                           & \underline{0.17}                        & 20,582                        &                         & 80.48 $\pm$ 0.34                        & 80.71 $\pm$ 1.85                        &                         & 93.76 $\pm$ 0.11                        & 83.41 $\pm$ 1.40                        &                         & 81.86 $\pm$ 0.34                        & 63.30 $\pm$ 1.25                        &                         & 99.44 $\pm$ 0.14                        & 91.97 $\pm$ 0.33                        &                         & 84.37                                      & 84.31                                     & 63.99                                     \\
        Attention-tuning \cite{Touvron2022}                        & 33.8                        & 20,792                        &                         & 82.38 $\pm$ 0.93                        & 77.11 $\pm$ 3.29                        &                         & 92.60 $\pm$ 0.74                        & 76.32 $\pm$ 4.05                        &                         & 75.58 $\pm$ 8.26                        & 51.31 $\pm$ 2.63                        &                         & 99.57 $\pm$ 0.14                        & \textbf{95.03 $\pm$ 0.30}               &                         & 81.24                                      & 71.59                                     & 61.49                                     \\
        Adapter \cite{Chen2022}                                 & 2.03                        & 19,360                        &                         & \textbf{87.95 $\pm$ 0.54}               & \underline{84.99 $\pm$ 1.89}                  &                         & \textbf{96.02 $\pm$ 0.13}               & 82.30 $\pm$ 2.26                        &                         & 77.61 $\pm$ 8.86                        & 57.56 $\pm$ 1.48                        &                         & 99.79 $\pm$ 0.12                        & 89.73 $\pm$ 4.55                        &                         & 84.49                                      & 83.76                                     & 64.88                                     \\
        BitFit \cite{Elad2022}                                  & \textbf{0.12}                        & 20,382                        &                         & 86.35 $\pm$ 0.42                        & 84.68 $\pm$ 0.42                        &                         & 95.11 $\pm$ 0.15                        & \textbf{84.60 $\pm$ 1.13}               &                         & 84.27 $\pm$ 0.34                        & \underline{66.03 $\pm$ 1.53}                  &                         & \underline{99.82 $\pm$ 0.04}                  & 93.43 $\pm$ 0.64                        &                         & 86.79                                      & 86.74                                     & 65.96                                     \\
        LoRA \cite{Hu2022}                                    & 0.68                        & 20,970                        &                         & 85.14 $\pm$ 3.86                        & 84.35 $\pm$ 1.27                        &                         & \underline{95.97 $\pm$ 0.40}                  & 81.06 $\pm$ 6.82                        &                         & 77.77 $\pm$ 1.92                        & 60.59 $\pm$ 2.36                        &                         & \textbf{99.89 $\pm$ 0.03}               & \underline{95.02 $\pm$ 0.70}                  &                         & 84.97                                      & 84.72                                     & 64.20                                     \\
        AdaLoRA \cite{Zhang2023}                                & 0.52                        & 20,960                        &                         & 86.39 $\pm$ 0.37                        & 84.27 $\pm$ 0.66                        &                         & 95.79 $\pm$ 0.25                        & 83.74 $\pm$ 0.88                        &                         & 88.16 $\pm$ 0.35                        & 64.39 $\pm$ 0.14                        &                         & 99.64 $\pm$ 0.08                        & 93.29 $\pm$ 0.79                        &                         & \underline{86.96}                                & \underline{86.76}                               & 65.71                                     \\
        FPT \cite{Huang2024FPT}                                    & 1.81                        & \underline{1,824}                         &                         & 83.79 $\pm$ 0.76                        & 84.69 $\pm$ 1.06                        &                         & 94.27 $\pm$ 0.18                        & 83.05 $\pm$ 0.85                        &                         & \underline{90.08 $\pm$ 0.68}                  & \textbf{66.23 $\pm$ 0.46}               &                         & 99.72 $\pm$ 0.05                        & 91.55 $\pm$ 0.81                        &                         & 86.67                                      & 86.00                                     & \underline{83.86}                               \\
        FPT+ (Ours)                                    & 1.03                        & \textbf{736}                          &                         & \underline{86.52 $\pm$ 0.42}                  & \textbf{86.13 $\pm$ 0.85}               &                         & 93.82 $\pm$ 0.40                        & \underline{84.14 $\pm$ 1.65}                  &                         & \textbf{90.50 $\pm$ 0.37}               & 65.75 $\pm$ 0.44                        &                         & 99.73 $\pm$ 0.14                        & 91.52 $\pm$ 0.72                        &                         & \textbf{87.26}                             & \textbf{86.87}                            & \textbf{86.08}                            \\ \hline
        \multicolumn{19}{c}{\textit{BioMedCLIP}} \\ \hline
        {\color[HTML]{9B9B9B} Full fine-tuning} & {\color[HTML]{9B9B9B} 100}  & {\color[HTML]{9B9B9B} 23,128} & & {\color[HTML]{9B9B9B} 84.40 $\pm$ 0.60} & {\color[HTML]{9B9B9B} 83.23 $\pm$ 0.94} & & {\color[HTML]{9B9B9B} 96.52 $\pm$ 0.02} & {\color[HTML]{9B9B9B} 81.61 $\pm$ 0.80} & & {\color[HTML]{9B9B9B} 88.05 $\pm$ 0.44} & {\color[HTML]{9B9B9B} 63.56 $\pm$ 2.46} & & {\color[HTML]{9B9B9B} 99.80 $\pm$ 0.04} & {\color[HTML]{9B9B9B} 96.87 $\pm$ 0.32} & & {\color[HTML]{9B9B9B} 86.52} & {\color[HTML]{9B9B9B} 64.03} & {\color[HTML]{9B9B9B} 64.03} \\
        {\color[HTML]{9B9B9B} Linear probing} & {\color[HTML]{9B9B9B} 0.01} & {\color[HTML]{9B9B9B} 3,416} & & {\color[HTML]{9B9B9B} 78.98 $\pm$ 0.28} & {\color[HTML]{9B9B9B} 77.38 $\pm$ 0.78} & & {\color[HTML]{9B9B9B} 93.65 $\pm$ 0.04} & {\color[HTML]{9B9B9B} 75.68 $\pm$ 1.76} & & {\color[HTML]{9B9B9B} 78.42 $\pm$ 0.13} & {\color[HTML]{9B9B9B} 62.89 $\pm$ 0.29} & & {\color[HTML]{9B9B9B} 99.50 $\pm$ 0.05} & {\color[HTML]{9B9B9B} 92.13 $\pm$ 0.51} & & {\color[HTML]{9B9B9B} 82.23} & {\color[HTML]{9B9B9B} 82.23} & {\color[HTML]{9B9B9B} 77.45} \\
        Prompt-tuning \cite{Jia2022} & \underline{0.17}                        & 20,582 & & 79.36 $\pm$ 0.36 & 77.22 $\pm$ 0.91 & & 93.84 $\pm$ 0.10 & 76.49 $\pm$ 0.69 & & 78.94 $\pm$ 0.12 & 62.24 $\pm$ 0.79 & & 99.56 $\pm$ 0.03 & 92.05 $\pm$ 0.45 & & 82.39 & 82.33 & 62.49 \\
        Attention-tuning \cite{Touvron2022}                        & 33.8                        & 20,792  & & 80.46 $\pm$ 1.67 & 77.58 $\pm$ 2.50 & & 94.28 $\pm$ 0.65 & 77.79 $\pm$ 3.24 & & 79.09 $\pm$ 0.36 & 60.19 $\pm$ 4.38 & & 99.72 $\pm$ 0.06 & 92.12 $\pm$ 2.08 & & 83.24 & 73.35 & 63.00 \\
        Adapter \cite{Chen2022}                                 & 2.03                        & 19,360  & & \textbf{84.16 $\pm$ 0.62} & 78.15 $\pm$ 0.60 & & \underline{95.69 $\pm$ 0.28} & 79.17 $\pm$ 2.39 & & 81.63 $\pm$ 2.19 & 61.83 $\pm$ 1.54 & & 99.56 $\pm$ 0.25 & 90.32 $\pm$ 4.47 & & 83.71 & 82.98 & 64.28 \\
        BitFit \cite{Elad2022}                                  & \textbf{0.12}                        & 20,382  & & 81.78 $\pm$ 0.72 & \underline{82.73 $\pm$ 1.14} & & 94.92 $\pm$ 0.10 & 82.20 $\pm$ 1.06 & & 80.78 $\pm$ 0.41 & \underline{65.17 $\pm$ 1.29} & & 99.61 $\pm$ 0.12 & \textbf{94.92 $\pm$ 0.98} & & 84.87 & 84.83 & 64.50 \\
        LoRA \cite{Hu2022}                                    & 0.68                        & 20,970   & & 81.65 $\pm$ 0.53 & 79.91 $\pm$ 0.46 & & \textbf{95.94 $\pm$ 0.22} & 76.82 $\pm$ 2.83 & & 81.01 $\pm$ 0.59 & 62.43 $\pm$ 1.63 & & \underline{99.78 $\pm$ 0.03} & 94.59 $\pm$ 1.49 & & 83.67 & 83.42 & 63.22 \\
        AdaLoRA \cite{Zhang2023}                                & 0.52                        & 20,960  & & \underline{83.78 $\pm$ 0.54} & 80.73 $\pm$ 1.39 & & 95.41 $\pm$ 0.22 & 80.07 $\pm$ 0.99 & & 84.03 $\pm$ 0.25 & 64.01 $\pm$ 1.00 & & 99.72 $\pm$ 0.02 & \underline{94.66 $\pm$ 0.89} & & 85.24 & \underline{85.05} & 64.41 \\
        FPT \cite{Huang2024FPT}                                    & 1.81                        & \underline{1,824}  & & 82.48 $\pm$ 0.36 & 79.81 $\pm$ 2.12 & & 94.33 $\pm$ 0.64 & \underline{83.82 $\pm$ 0.72} & & \underline{86.91 $\pm$ 0.43} & 64.06 $\pm$ 0.91 & & 99.67 $\pm$ 0.08 & 92.26 $\pm$ 0.76 & & \underline{85.30} & 84.64 & \underline{82.53} \\
        FPT+ (Ours)                                    & 1.03                        & \textbf{736} & & 82.38 $\pm$ 0.25 & \textbf{83.73 $\pm$ 0.30} & & 94.50 $\pm$ 0.31 & \textbf{83.88 $\pm$ 0.83} & & \textbf{88.53 $\pm$ 0.97} & \textbf{65.98 $\pm$ 0.82} & & \textbf{99.84 $\pm$ 0.06} & 94.57 $\pm$ 0.14 & & \textbf{86.81} & \textbf{86.42} & \textbf{85.64} \\
        \bottomrule
        \end{tabular}
    }
    \label{main}
\end{table*}

\begin{table}[t]
    \centering
    \renewcommand{\arraystretch}{1.2}
    \caption{Training efficiency comparisons across different PETL methods on the Messidor2 dataset.}
    \resizebox{\columnwidth}{!}{%
        \begin{tabular}{lccccc}
        \toprule
                                          & FLOPs (GB) $\downarrow$                   & \begin{tabular}[c]{@{}c@{}}Total Training\\ time (s)\end{tabular} $\downarrow$ &  \begin{tabular}[c]{@{}c@{}}Throughput\\ (img/s)\end{tabular} $\uparrow$ & \begin{tabular}[c]{@{}c@{}}Peak Power\\ Usage (W)\end{tabular} $\downarrow$ & \begin{tabular}[c]{@{}c@{}}Avg. Power\\ Usage (W)\end{tabular} $\downarrow$ \\ \hline
        {\color[HTML]{9B9B9B} Full FT}    & {\color[HTML]{9B9B9B} 87.84} & {\color[HTML]{9B9B9B} 1,135.91}                                       & {\color[HTML]{9B9B9B} 18.48}                                 & {\color[HTML]{9B9B9B} 295.62}                                  & {\color[HTML]{9B9B9B} 280.92}                                  \\
        {\color[HTML]{9B9B9B} Linear}     & {\color[HTML]{9B9B9B} 87.84} & {\color[HTML]{9B9B9B} 416.72}                                         & {\color[HTML]{9B9B9B} 51.06}                                 & {\color[HTML]{9B9B9B} 286.61}                                  & {\color[HTML]{9B9B9B} 278.59}                                  \\
        Prompt-tuning \cite{Jia2022}                            & 88.30                        & 946.55                                                                & 22.19                                                        & 296.69                                                         & 286.52                                                         \\
        Attention-tuning \cite{Touvron2022}                         & 87.84                        & 1,005.62                                                                & 20.89                                                        & 291.76                                                         & 281.79                                                         \\
        Adapter \cite{Chen2022}                           & 89.68                        & 954.51                                                                & 21.93                                                        & 297.30                                                         & 288.30                                                         \\
        BitFit \cite{Elad2022}                              & 87.84                        & 966.69                                                                & 21.75                                                        & 291.40                                                         & 279.66                                                         \\
        LoRA \cite{Hu2022}                              & 107.89                       & 1,014.68                                                                & 20.62                                                        & 295.85                                                         & 288.47                                                         \\
        AdaLoRA \cite{Zhang2023}                           & 107.28                       & 1,027.17                                                                & 20.44                                                        & 292.00                                                         & 280.56                                                         \\ \hline
        {\color[HTML]{9B9B9B} Preloading} & {\color[HTML]{9B9B9B} 87.84} & {\color[HTML]{9B9B9B} 31.76}                                              & {\color[HTML]{9B9B9B} 32.81}                                 & {\color[HTML]{9B9B9B} 221.81}                                 & {\color[HTML]{9B9B9B} 217.74}                                  \\
        FPT \cite{Huang2024FPT}                              & \underline{0.28}             & \underline{154.23}                                                     & \underline{171.83}                                           & \underline{166.54}                                             & \underline{183.10}                                             \\
        FPT+                              & \textbf{0.06}                & \textbf{90.61}                                                        & \textbf{357.37}                                              & \textbf{114.38}                                                & \textbf{107.60}                                                \\ \bottomrule
        \end{tabular}
}
    \label{efficency}
\end{table}

\begin{table*}[t]
    \centering
    \renewcommand{\arraystretch}{1.2}
    \caption{Comparison results with state-of-the-art PETL methods with the ImageNet-based supervised LPM, evaluated at different input resolutions. The '/' symbol indicates out-of-memory conditions on a 48GB GPU. The average AUC is computed across all eight classification tasks.}
    \resizebox{0.68\textwidth}{!}{%
        \begin{tabular}{lccccccccccc}
        \toprule
                                              & \multicolumn{5}{c}{Memory Usage $\downarrow$}                                                                                                                      &                         & \multicolumn{5}{c}{Average AUC $\uparrow$}                                                                                                                        \\ \cline{2-6} \cline{8-12} 
        \multirow{-2}{*}{Method}              & 128                         & 256                          & 384                          & 512                         & 640                         &                         & 128                          & 256                          & 384                          & 512                          & 640                        \\ \hline
        \multicolumn{12}{c}{\textit{ViT-B}} \\ \hline
        {\color[HTML]{9B9B9B} Full FT}         & {\color[HTML]{9B9B9B} 2,716} & {\color[HTML]{9B9B9B} 4,540} & {\color[HTML]{9B9B9B} 10,034} & {\color[HTML]{9B9B9B} 23,128} & {\color[HTML]{9B9B9B} 48,268} & & {\color[HTML]{9B9B9B} 83.63} & {\color[HTML]{9B9B9B} 86.28} & {\color[HTML]{9B9B9B} 87.50} & {\color[HTML]{9B9B9B} 88.82} & {\color[HTML]{9B9B9B} 89.11} \\
        {\color[HTML]{9B9B9B} Linear probing}     & {\color[HTML]{9B9B9B} 842} & {\color[HTML]{9B9B9B} 1,416} & {\color[HTML]{9B9B9B} 1,668} & {\color[HTML]{9B9B9B} 3,416} & {\color[HTML]{9B9B9B} 6,836} & & {\color[HTML]{9B9B9B} 77.81} & {\color[HTML]{9B9B9B} 82.06} & {\color[HTML]{9B9B9B} 82.69} & {\color[HTML]{9B9B9B} 83.28} & {\color[HTML]{9B9B9B} 82.62} \\
        Prompt-tuning \cite{Jia2022}    & 1,266 & 3,094 & 8,204 & 20,582 & 45,352 & & 80.03 & 82.68 & 83.62 & 83.69 & 83.33 \\
        Attention-tuning \cite{Touvron2022} & 1,584 & 3,492 & 8,300 & 20,792 & 45,396 & & 79.98 & 82.51 & 82.94 & 83.68 & 82.35 \\
        Adapter \cite{Chen2022}   & 1,280 & 3,018 & 7,962 & 19,360 & 43,516 & & 80.94 & 81.61 & 82.80 & 84.14 & 84.69 \\
        BitFit \cite{Elad2022}    & 1,250 & 2,964 & 8,072 & 20,382 & 45,392 & & 80.80 & 84.87 & 85.34 & 85.67 & 86.96 \\
        LoRA \cite{Hu2022}      & 1,294 & 3,132 & 8,472 & 20,970 & 46,554 & &\textbf{81.93} & 84.67 & 85.48 & 85.94 & 86.96 \\
        AdaLoRA \cite{Zhang2023}   & 1,290 & 3,120 & 8,360 & 20,960 & 45,798 & & 81.10 & 84.86 & \underline{85.97} & \underline{86.59} & \underline{87.11} \\
        FPT \cite{Huang2024FPT}       & \underline{1,430} & \underline{1,522} & \underline{1,682} & \underline{1,824} & \underline{2,166} & & 81.22 & \underline{84.92} & 85.63 & 86.40 & 86.19 \\
        FPT+ (Ours)      & \textbf{549} & \textbf{624} & \textbf{698} & \textbf{736} & \textbf{896} & & \underline{81.29} & \textbf{85.07} & \textbf{86.02} & \textbf{87.12} & \textbf{87.54} \\ \hline
        \multicolumn{12}{c}{\textit{ViT-L}} \\ \hline
        
        {\color[HTML]{9B9B9B} Full FT}        & {\color[HTML]{9B9B9B} 7,816} & {\color[HTML]{9B9B9B} 12,168} & {\color[HTML]{9B9B9B} 26,470} & {\color[HTML]{9B9B9B} /}    & {\color[HTML]{9B9B9B} /}    & {\color[HTML]{9B9B9B} } & {\color[HTML]{9B9B9B} 83.94} & {\color[HTML]{9B9B9B} 86.96} & {\color[HTML]{9B9B9B} 87.23} & {\color[HTML]{9B9B9B} /}     & {\color[HTML]{9B9B9B} /}   \\
        {\color[HTML]{9B9B9B} Linear probing} & {\color[HTML]{9B9B9B} 1,640} & {\color[HTML]{9B9B9B} 1,884}  & {\color[HTML]{9B9B9B} 2,778}  & {\color[HTML]{9B9B9B} 5,066} & {\color[HTML]{9B9B9B} 9,716} & {\color[HTML]{9B9B9B} } & {\color[HTML]{9B9B9B} 77.46} & {\color[HTML]{9B9B9B} 82.99} & {\color[HTML]{9B9B9B} 84.38} & {\color[HTML]{9B9B9B} 84.62} & {\color[HTML]{9B9B9B} 84.77} \\
        Prompt-tuning \cite{Jia2022}                        & 3,128                        & 7,602                         & 20,392                        & /                           & /                           &                         & 79.82                         & 83.90                             & 84.95                             & /                            & /                          \\
        Attention-tuning \cite{Touvron2022}                     & 4,340                        & 8,710                         & 21,516                        & /                           & /                           &                         & 80.93                        & 82.21                        & 83.89                        & /                            & /                          \\
        Adapter \cite{Chen2022}                              & 3,248                        & 7,608                         & 20,394                        & /                           & /                           &                         & 79.30                         & 81.70                         & 82.88                        & /                            & /                          \\
        BitFit \cite{Elad2022}                               & 3,058                        & 7,186                         & 20,074                        & /                           & /                           &                         & \textbf{82.85}                        & 85.46                        & 86.38                        & /                            & /                          \\
        LoRA \cite{Hu2022}                                 & 3,152                        & 7,810                         & 21,190                        & /                           & /                           &                         & 80.13                        & 82.96                        & 83.95                        & /                            & /                          \\
        AdaLoRA \cite{Zhang2023}                              & 3,142                        & 7,806                         & 21,056                        & /                           & /                           &                         & \underline{82.17}                        & \underline{85.52}                        & \underline{86.46}                        & /                            & /                          \\
        FPT \cite{Huang2024FPT}                                  & \underline{2,546}                             & \underline{2,724}                             & \underline{3,152}                             & \underline{3,730}                             & \underline{4,456}             &          & 80.72                            & 84.34                             & 85.75                             & \underline{86.85}                            & \underline{87.28}                                                                      \\
        FPT+ (Ours)                           & \textbf{570}                         & \textbf{630}                          & \textbf{718}                         & \textbf{862}                        & \textbf{1,050}                        &                         & 81.74                        & \textbf{85.60}                        & \textbf{86.88}                        & \textbf{87.87}                        & \textbf{88.14}                      \\ \bottomrule
        \end{tabular}
    }
    \label{vitl_resolution}
\end{table*}

\subsection{Comparisons with State-of-the-art}
We compare FPT+ against full fine-tuning, linear probing, and SOTA PETL approaches. In full fine-tuning, all parameters are learnable when training on downstream tasks. Although full fine-tuning often represents the performance's upper bound for fine-tuning, the relative scarcity of training samples in medical image datasets may lead to overfitting and consequent performance degradation. Linear probing emerges as an efficient strategy to adapt pre-trained models to downstream tasks by solely training a new task-specific head on top of the pre-trained model. Additionally, we compare FPT+ with popular SOTA PETL methods applicable to vision tasks.

The performance comparisons are tabulated in Table~\ref{main}. Full fine-tuning presents strong performance across all datasets but comes with the highest computational cost, requiring a 100\% parameter ratio and significant memory usage (23,128 MB), making it impractical for resource-constrained settings. Linear probing, on the other hand, shows low parameter usage (0.01\%) and memory consumption (3,416 MB). However, this method generally underperforms. Other PETL methods are more balanced between parameter efficiency and performance. For example, LoRA and AdaLoRA, with 0.68\% and 0.52\% parameter ratios respectively, show competitive performance across various datasets.

Although these aforementioned methods achieve reductions in the number of learnable parameters, their memory consumption remains high. For instance, Adapter has the lowest memory consumption but still requires 19,360 MB to fine-tune the LPM. In contrast, our proposed method, FPT+, outperforms other PETL methods in both performance and efficiency, achieving the best performance among PETL methods and the best performance-efficiency balance (PPE and PME) among all methods across different pre-trained weights. Specifically, FPT+ exhibits remarkable parameter efficiency with only 1.03\% of learnable parameters and significantly lower memory consumption (736 MB, 3.18\% compared to full fine-tuning), without compromising performance. FPT already shows significant improvements in parameter efficiency and performance. On top of FPT, FPT+ further enhances these metrics, achieving higher performance and reduced memory consumption (736 MB vs. 1,824 MB). Notably, the memory consumption reported for FPT is lower than that reported in its original paper, due to an optimized preloading mechanism in FPT+ that stores only the important tokens from the frozen LPM. This optimization, applied to both FPT+ and FPT, reduces memory usage without affecting the AUC performance.

In addition to using ImageNet-based pre-trained weights, we evaluate FPT+ with BioMedCLIP \cite{Zhang2023biomedclip}, a widely used vision-language model pre-trained on a diverse set of medical imaging modalities. In this evaluation, the vision encoder of BioMedCLIP is adopted as the LPM. As shown in Table~\ref{main}, FPT+ with BioMedCLIP consistently outperforms other PETL methods across five datasets, achieving the highest average AUC. Notably, it surpasses the full fine-tuning baseline in terms of the average AUC, further highlighting its ability to effectively leverage medical-specific pre-trained knowledge.

The comparison results emphasize the effectiveness of FPT+ in balancing performance, parameter efficiency, and memory consumption. Its design ensures efficient high-resolution medical image classification, making it a promising solution for advancing medical diagnostics. The generalizability and robustness of FPT+ are further highlighted by its effectiveness with different pre-trained weights, making it adaptable to different pre-training paradigms and practical for a wide range of medical image classification tasks.

\subsection{Training Efficiency}
To comprehensively evaluate the efficiency of PETL methods, we further assess several metrics related to training cost, including computational cost (FLOPs), total training time, training time per epoch, throughput, and power usage (both peak and average). These training efficiency metrics are evaluated using the Messidor2 dataset, containing 1,748 samples with a resolution of 512 × 512, and employing ViT-B as the backbone. Table~\ref{efficency} provides a detailed comparison of these metrics across various methods.

As shown in Table~\ref{efficency}, FPT+ demonstrates remarkable training efficiency with computational cost of only 0.06 GB FLOPs, being significantly lower than full fine-tuning (87.84 GB) and PETL methods like LoRA (107.89 GB). FPT+ also excels in terms of throughput, processing 357.37 images per second, which is a significant improvement over other PETL methods like LoRA (20.62 images/second) and Prompt-tuning (22.19 images/second). Compared to FPT, FPT+ doubles the throughput, showing its enhanced efficiency. With these reductions in the computational cost, FPT+ achieves a total training time of 90.61 seconds, which is more than 12 times faster than full fine-tuning (1,135.91 seconds). FPT+ is also energy-efficient, achieving an average power usage of 107.60 W and a peak power usage of 114.38 W, staying far below the average of 280 W and the peak usage of over 290 W for other PETL methods on GPUs with a 300 W power limit.

Additionally, FPT+ incorporates a preloading step, which extracts and pre-stores features from the frozen LPM before training. This step, taking only 31.76 seconds, adds negligible overhead compared to the substantial training costs of other fine-tuning methods.

These results establish FPT+ as a highly efficient PETL method in terms of computational, temporal, and energy resources, making it a practical and effective solution for high-resolution medical image classification tasks, especially in resource-constrained settings.

\subsection{Evaluation with Different Resolutions}
To further validate the resolution scalability and memory efficiency of FPT+, we compare its performance against other PETL baselines using the ViT-B backbone with ImageNet-based supervised LPM across five input resolutions: 128 × 128, 256 × 256, 384 × 384, 512 × 512, and 640 × 640.

As shown in the top panel of Table~\ref{vitl_resolution}, FPT+ consistently achieves strong AUC performance across all resolutions. Notably, it achieves the best average AUC across all resolutions except for the lowest resolution (128 × 128), where it still demonstrates competitive performance compared to other PETL methods. Importantly, FPT+ demonstrates exceptional memory efficiency. At a resolution of 640 × 640, FPT+ consumes only 896 MB, which is significantly lower than all competing PETL methods (e.g., 46,554 MB for LoRA). This highlights the scalability of FPT+ in handling high-resolution inputs with minimal computational overhead.

These experimental results also confirm that FPT+ not only scales effectively across a wide range of resolutions but also achieves an optimal balance between performance and efficiency, making it well-suited for deployment in various clinical settings.

\begin{table*}[t]
    \centering
    \renewcommand{\arraystretch}{1.2}
    \caption{Performance of FPT+ using ViT-B with a patch size of 8 and the ImageNet-based supervised LPM at a resolution of 256 × 256.}
    \resizebox{\textwidth}{!}{%
        \begin{tabular}{lcccccccccccccccccc}
        \toprule
                                                & \multicolumn{2}{c}{Computing cost}                         &                         & \multicolumn{2}{c}{Fundus image}                                          & \multicolumn{1}{c}{}    & \multicolumn{2}{c}{Dermoscopic image}                                     & \multicolumn{1}{c}{}    & \multicolumn{2}{c}{Mammography}                                           & \multicolumn{1}{c}{}    & \multicolumn{2}{c}{Chest X-ray}                                           &                         & \multicolumn{1}{c}{}                       & \multicolumn{1}{c}{}                      & \multicolumn{1}{c}{}                      \\ \cline{2-3} \cline{5-6} \cline{8-9} \cline{11-12} \cline{14-15}
        \multirow{-2}{*}{Method}                & Params. $\downarrow$                     & Mem. $\downarrow$                         &                         & \multicolumn{1}{c}{DDR}             & \multicolumn{1}{c}{Messidor-2}       & \multicolumn{1}{c}{}    & \multicolumn{1}{c}{ISIC2018}        & \multicolumn{1}{c}{ISIC2016}        & \multicolumn{1}{c}{}    & \multicolumn{1}{c}{DDSM}            & \multicolumn{1}{c}{CMMD}            & \multicolumn{1}{c}{}    & \multicolumn{1}{c}{COVID}           & \multicolumn{1}{c}{CHNCXR}              &                         & \multicolumn{1}{c}{\multirow{-2}{*}{Avg. $\uparrow$}} & \multicolumn{1}{c}{\multirow{-2}{*}{PPE $\uparrow$}} & \multicolumn{1}{c}{\multirow{-2}{*}{PME $\uparrow$}} \\ \hline

        {\color[HTML]{9B9B9B} Full FT}     & {\color[HTML]{9B9B9B} 100}  & {\color[HTML]{9B9B9B} 23,078} & {\color[HTML]{9B9B9B} } & {\color[HTML]{9B9B9B} 87.76 $\pm$ 0.79} & {\color[HTML]{9B9B9B} 84.95 $\pm$ 0.99} & {\color[HTML]{9B9B9B} } & {\color[HTML]{9B9B9B} 96.52 $\pm$ 0.06} & {\color[HTML]{9B9B9B} 81.87 $\pm$ 0.69} & {\color[HTML]{9B9B9B} } & {\color[HTML]{9B9B9B} 93.18 $\pm$ 0.45} & {\color[HTML]{9B9B9B} 67.26 $\pm$ 1.76} & {\color[HTML]{9B9B9B} } & {\color[HTML]{9B9B9B} 99.85 $\pm$ 0.03} & {\color[HTML]{9B9B9B} 96.27 $\pm$ 0.36} & {\color[HTML]{9B9B9B} } & {\color[HTML]{9B9B9B} 88.46} & {\color[HTML]{9B9B9B} 65.47} & {\color[HTML]{9B9B9B} 65.47} \\
        {\color[HTML]{9B9B9B} Linear probing} & {\color[HTML]{9B9B9B} 0.01} & {\color[HTML]{9B9B9B} 3,366}  & {\color[HTML]{9B9B9B} } & {\color[HTML]{9B9B9B} 81.55 $\pm$ 0.48} & {\color[HTML]{9B9B9B} 82.34 $\pm$ 0.36} & {\color[HTML]{9B9B9B} } & {\color[HTML]{9B9B9B} 93.65 $\pm$ 0.04} & {\color[HTML]{9B9B9B} 80.94 $\pm$ 0.62} & {\color[HTML]{9B9B9B} } & {\color[HTML]{9B9B9B} 81.94 $\pm$ 0.34} & {\color[HTML]{9B9B9B} 59.89 $\pm$ 1.54} & {\color[HTML]{9B9B9B} } & {\color[HTML]{9B9B9B} 99.23 $\pm$ 0.11} & {\color[HTML]{9B9B9B} 90.11 $\pm$ 0.42} & {\color[HTML]{9B9B9B} } & {\color[HTML]{9B9B9B} 83.71} & {\color[HTML]{9B9B9B} 83.71} & {\color[HTML]{9B9B9B} 78.90} \\
        Prompt-tuning \cite{Jia2022}                        & \underline{0.17}                  & 20,530                        &                         & 81.72 $\pm$ 0.56                        & 82.26 $\pm$ 0.31                        &                         & 93.84 $\pm$ 0.10                        & 81.31 $\pm$ 0.70                        &                         & 83.03 $\pm$ 0.32                        & 59.83 $\pm$ 1.01                        &                         & 99.25 $\pm$ 0.09                        & 92.78 $\pm$ 0.30                        &                         & 84.25                        & 84.19                        & 63.91                        \\
        Attention-tuning \cite{Touvron2022}                     & 32.97                       & 20,740                        &                         & 83.28 $\pm$ 1.46                        & 82.92 $\pm$ 1.30                        &                         & 94.28 $\pm$ 0.65                        & 78.39 $\pm$ 2.25                        &                         & 82.94 $\pm$ 3.06                        & 60.17 $\pm$ 5.15                        &                         & \underline{99.79 $\pm$ 0.06}                  & 92.21 $\pm$ 0.67                        &                         & 84.25                        & 74.44                        & 63.77                        \\
        Adapter \cite{Chen2022}                       & 2.04                        & 19,596                        &                         & \textbf{88.33 $\pm$ 0.82}               & 81.69 $\pm$ 1.68                        &                         & 93.69 $\pm$ 0.28                        & 78.39 $\pm$ 3.09                        &                         & 81.81 $\pm$ 3.24                        & 61.19 $\pm$ 0.44                        &                         & \textbf{99.84 $\pm$ 0.03}               & 89.96 $\pm$ 1.85                        &                         & 84.36                        & 83.62                        & 64.59                        \\
        BitFit \cite{Elad2022}                        & \textbf{0.12}               & 20,330                        &                         & 85.02 $\pm$ 0.24                        & 84.69 $\pm$ 0.32                        &                         & 93.92 $\pm$ 0.10                        & 82.20 $\pm$ 1.60                        &                         & 83.44 $\pm$ 0.26                        & \underline{65.17 $\pm$ 1.54}                  &                         & 99.68 $\pm$ 0.07                        & 92.57 $\pm$ 0.67                        &                         & 85.84                        & 85.80                         & 65.24                        \\
        LoRA \cite{Hu2022}                          & 0.69                        & 20,936                        &                         & \underline{86.38 $\pm$ 0.47}                  & \underline{85.03 $\pm$ 0.83}                  &                         & \underline{94.44 $\pm$ 0.22}                  & 79.96 $\pm$ 3.40                        &                         & 80.34 $\pm$ 2.59                        & 59.78 $\pm$ 1.08                        &                         & \textbf{99.84 $\pm$ 0.04}               & \textbf{93.77 $\pm$ 0.76}               &                         & 84.94                        & 84.69                        & 64.17                        \\
        AdaLoRA \cite{Zhang2023}                       & 0.52                        & 20,926                        &                         & 86.24 $\pm$ 0.35                        & 84.78 $\pm$ 0.55                        &                         & 94.41 $\pm$ 0.22                        & \underline{82.95 $\pm$ 1.14}                  &                         & 87.13 $\pm$ 0.50                        & 64.11 $\pm$ 0.88                        &                         & 99.50 $\pm$ 0.12                        & \underline{93.50 $\pm$ 0.52}                  &                         & \underline{86.58}                  & \underline{86.39}                  & 65.42                        \\
        FPT \cite{Huang2024FPT}                           & 1.83                        & \underline{9,144}                   &                         & 84.07 $\pm$ 0.54                        & 84.27 $\pm$ 0.89                        &                         & 94.33 $\pm$ 0.64                        & 82.16 $\pm$ 1.82                        &                         & \underline{89.35 $\pm$ 0.74}                  & 63.85 $\pm$ 0.52                        &                         & 98.77 $\pm$ 0.41                        & 89.35 $\pm$ 2.85                        &                         & 85.77                        & 85.10                        & \underline{74.20}                  \\
        FPT+ (Ours)                          & 0.99                        & \textbf{1,494}                &                         & 85.48 $\pm$ 0.74                        & \textbf{86.95 $\pm$ 0.41}               &                         & \textbf{94.50 $\pm$ 0.30}               & \textbf{83.93 $\pm$ 1.06}               &                         & \textbf{89.85 $\pm$ 1.25}               & \textbf{67.46 $\pm$ 0.45}               &                         & 99.54 $\pm$ 0.20                        & 90.28 $\pm$ 2.24                        &                         & \textbf{87.25}               & \textbf{86.88}               & \textbf{84.91}    \\
        \bottomrule
        \end{tabular}
    }
    \label{vitb_patch8}
\end{table*}

\subsection{Evaluation with A Larger Pre-trained Model}
To further assess the generalizability of FPT+ and highlight its significant advantage in memory saving when scaling model capacity, we evaluate its performance using the ViT-L backbone across various input resolutions. As shown in Table~\ref{vitl_resolution}, most PETL methods encounter out-of-memory issues on a 48GB GPU when the input resolution exceeds 384 $\times$ 384, limiting their capability to improve performance through higher resolutions. In contrast, FPT, FPT+, and linear probing, with better memory efficiency, can run at higher resolutions and exhibit greater performance potential.

Full fine-tuning achieves its best performance with an average AUC of 87.23, at its highest feasible resolution of 384 $\times$ 384. Similarly, the PETL method AdaLoRA achieves its best performance with an average AUC of 86.48, also at the resolution of 384 $\times$ 384. These methods are constrained by memory limitations, which largely restrict their feasibility to operate at higher resolutions. Although linear probing can run at a resolution of 640 $\times$ 640, its performance remains relatively low, with an average AUC of 84.77 at the resolution of 640 $\times$ 640. In contrast, FPT+ not only achieves the best performance among all PETL methods at the resolution of 384 $\times$ 384 with an average AUC of 86.88 but also runs efficiently at the higher resolution of 640 $\times$ 640, achieving the highest overall performance with an average AUC of 88.14. These results demonstrate the superior scalability of FPT+ when handling higher resolutions.

One of the key advantages of FPT+ is its exceptional memory efficiency. At a resolution of 640 $\times$ 640, FPT+ has a memory usage of 1,050 MB, which is even lower than the memory usage of other methods at a resolution of 128 $\times$ 128. Such memory efficiency enhances the feasibility of deploying FPT+ and enables the exploration of higher input resolutions to potentially attain even better performance.

\begin{figure*}[t]
	\centering
	\includegraphics[width=\textwidth]{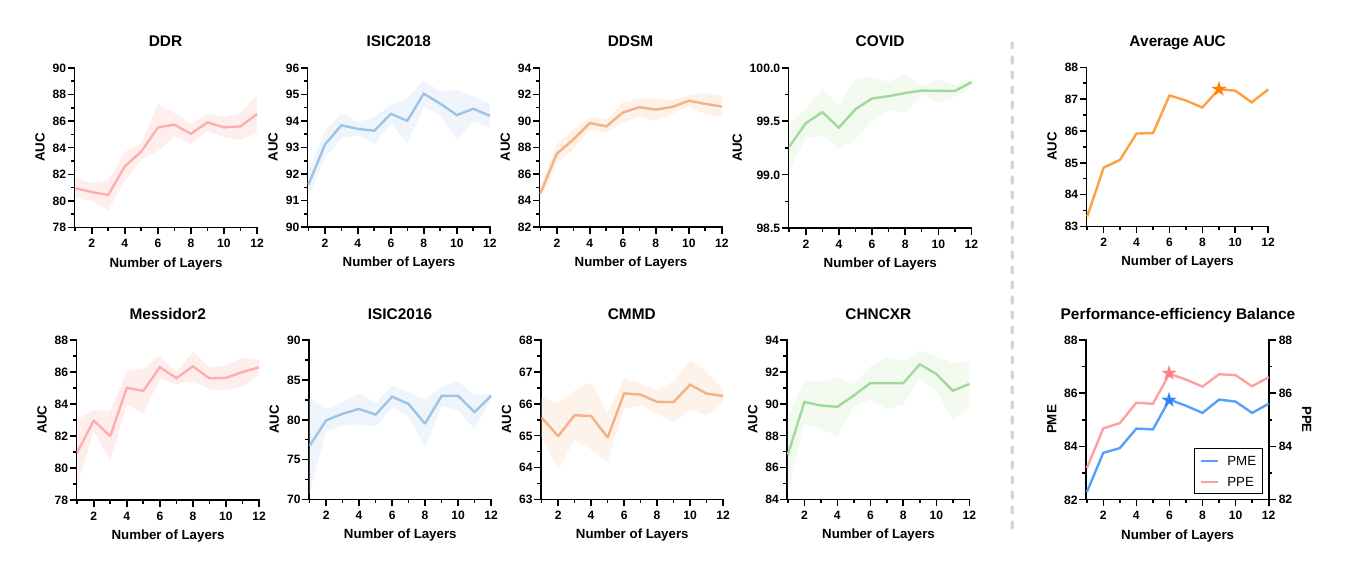}
	\caption{Impact of the number of layers on FPT+'s performance and efficiency. The star symbol indicates the best performance.}
	\label{num_layer}
\end{figure*}

\begin{figure}[t]
	\centering
	\includegraphics[width=0.65\columnwidth]{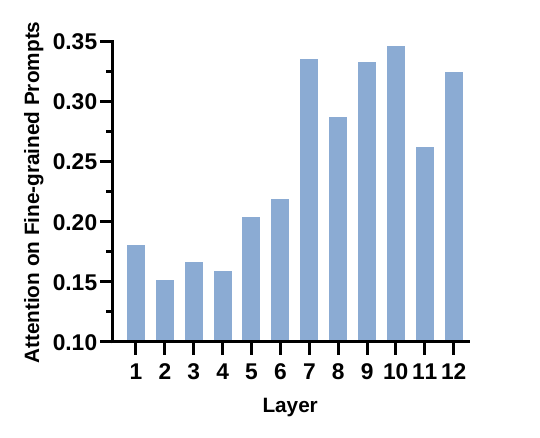}
	\caption{Average attention score on the fine-grained prompts at each layer.}
	\label{attn_on_prompts}
\end{figure}

\subsection{Evaluation with Smaller Patch Size}
To further explore the adaptability of FPT+ to different ViT configurations, we evaluate its performance using ViT-B with a smaller patch size of 8 at an input resolution of 256 × 256. Although a smaller patch size can help the model capture fine-grained information, it substantially increases the number of input tokens, leading to significantly higher computational costs. As only FPT+, FPT, and linear probing are feasible at a resolution of 512 × 512 on a 48GB GPU under this configuration, We conduct experiments at a resolution of 256 × 256 for all methods to enable a fair comparison.

As shown in Table~\ref{vitb_patch8}, FPT+ maintains remarkable efficiency under this memory-demanding setup. While most PETL baselines consume approximately 20,000 MB of memory, FPT+ requires only 1,494 MB, a footprint comparable to that of ViT-B with a patch size of 16 at a 512 × 512 resolution. Despite its significantly lower memory usage, FPT+ achieves the best AUC performance on five out of eight evaluated datasets, along with the highest average AUC and performance-efficiency balance among all methods. These results further confirm the robustness of FPT+ across various ViT configurations.

\subsection{Contribution of Each Layer}
In this section, we analyze the behavior of FPT, in terms of fine-grained prompts and show how FPT+ empirically simplifies FPT's architecture. As described in FPT, similar to FPT+, the pre-trained knowledge is conveyed to the side network via fine-grained prompts and fusion modules. The fine-grained prompts are concatenated with the intermediate sequence of the side network to join the forward propagation. The side network integrates the pre-trained knowledge through self-attention layers, where other tokens obtain information from the fine-grained prompts by paying attention to these prompts. However, unlike FPT+, which has a side network of $L_S$ layers, in FPT, each layer in the LPM has a corresponding learnable layer and a corresponding fusion module, resulting in a side netowrk of $L_M$ layers, which induces non-trivial computational overhead.

Since higher layers of a ViT-based LPM are more likely to contain high-level information for image understanding \cite{raghu2021vision}, we conjecture that not all layers of the side network are necessary for high performance. To investigate the contribution of each layer to downstream tasks, we visualize the average attention score that other tokens pay to the fine-grained prompts at each layer of FPT. As shown in Fig.~\ref{attn_on_prompts}, the fusion modules pay more attention to the fine-grained prompts in higher layers; the average attention score of layers 7-12 is 0.315, while that of layers 1-6 is 0.180. This suggests that leveraging information from higher layers is more critical for downstream tasks, allowing for a more efficient design of the side network by focusing on key layers.

To determine the number of LPM layers necessary to provide sufficient information for downstream tasks, we sequentially add layers to the side network starting from the highest to the lowest, including the corresponding fusion modules, and assess their impact. As presented in the right part of Fig.~\ref{num_layer}, the average AUC increases with the number of layers and saturates at 6 layers, demonstrating that the pre-trained information from the highest 6 layers is sufficient for downstream tasks. Although FPT+ with 9 layers achieves the best average AUC, increasing the number of layers also increases the number of learnable parameters and memory consumption. Furthermore, FPT+ with 6 layers achieves very close AUC compared to 9 layers (87.12 vs 87.32), and the best performance-efficiency balance in terms of PPE and PME. Therefore, we set the default number of layers $L_S$ for the side network to 6.

\subsection{Ablation Study}
In this section, we systematically investigate the impact of each component in FPT+. Unless otherwise specified, the presented performance metric is the average AUC across eight datasets using the pre-trained weights in a supervised manner, and the configuration of FPT+ remains the same as that described in Section~\ref{setup}.

\subsubsection{Impact of Components}
\begin{figure}[t]
	\centering
	\includegraphics[width=0.85\columnwidth]{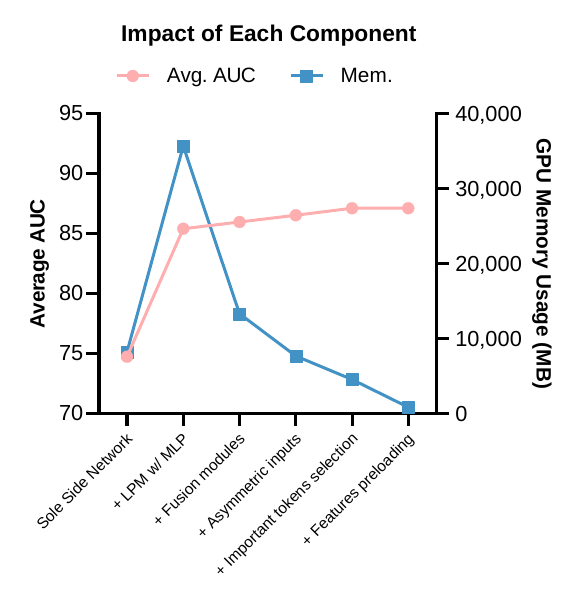}
	\caption{Impact of each component on FPT+'s performance and efficiency. The average AUC and GPU memory usage are shown for each configuration, starting with the side network alone and sequentially adding components.}
	\label{ablation}
\end{figure}

To assess the contributions of individual components within our proposed FPT+, we conduct an ablation study by adding components sequentially and measuring their impact on AUC and GPU memory usage. As shown in Fig.~\ref{ablation}, each additional component incrementally enhances the model's performance, with the average AUC improving from 74.74 to 87.12. The most substantial performance gain is attributed to the introduction of pre-trained knowledge from the LPM. However, simply using MLP for transferring pre-trained knowledge significantly increases the memory consumption from 8,214 MB to 35,682 MB. Replacing MLP with fusion modules notably enhances memory efficiency by reducing the memory consumption from 35,682 MB to 13,328 MB, followed by consistent improvements with the incorporation of asymmetric inputs, important tokens selection, and features preloading down to 736 MB. Notably, important tokens selection mechanism not only lowers memory consumption but also improves the AUC performance by reducing the number of non-informative tokens, allowing the fusion module to focus on diagnostically relevant features. By combining these components, FPT+ achieves the optimal balance of high performance and low memory consumption.

\begin{table}[t]
    \vspace{-4pt}
    \centering
    \renewcommand{\arraystretch}{1.2}
    \caption{FPT+'s performance and efficiency with different input resolutions for the side network.}
    \resizebox{0.68\columnwidth}{!}{%
        \begin{tabular}{cccc}
        \toprule
        Side Input Resolution & Avg. AUC $\uparrow$       & Mem. $\downarrow$         & PME $\uparrow$            \\ \hline
        64        & \underline{86.78}          & \textbf{694} & \underline{85.67}    \\
        128       & \textbf{87.12} & \underline{736}    & \textbf{85.94} \\
        256       & 86.32          & 1,342         & 84.23          \\
        512       & 84.78          & 8,704         & 73.80         \\ \bottomrule
        \end{tabular}
    }
    \label{resolution}
\end{table}

\subsubsection{Side Input Resolutions}
To reduce memory consumption, we propose asymmetric inputs for the LPM and the side network, where the input for the side network is a down-scaled version of that for the LPM. In this section, we investigate the impact of the input resolution for the side network. We train FPT+ using different input resolutions for the side network while maintaining the input resolution for the LPM at $512 \times 512$. As shown in Table~\ref{resolution}, the results indicate that a resolution of $128 \times 128$ yields optimal performance while maintaining memory efficiency. We conjecture that the poor performance observed with a higher down-scaled resolution could potentially be attributed to the lengthy input sequence in the side network, which distracts attention from the fine-grained prompts. Notably, even using an input resolution of $64 \times 64$ for the side network achieves better performance compared to other PETL methods, demonstrating that the proposed fine-grained prompts and fusion modules can effectively convey fine-grained information to the side network.

\subsubsection{Number of Fine-grained Prompts}
The number of fine-grained prompts can be significantly fewer than the total tokens from the input images presented in both the pre-trained model and the side network, thereby reducing the memory consumption. We here explore the influence of the number of fine-grained prompts. As depicted in Table~\ref{prompt}, with the increase in the number of prompts, the performance improves and the best performance is achieved at 16 prompts, which number is sufficient to summarize fine-grained information and convey it to the side network. We also notice that even with only 4 prompts, the performance is comparable to other PETL methods, demonstrating the effectiveness of FPT+'s fusion modules in obtaining pre-trained knowledge from the LPM.

\begin{table}[t]
    \vspace{-4pt}
    \centering
    \renewcommand{\arraystretch}{1.2}
    \caption{FPT+'s performance and efficiency with different numbers of prompts for the side network.}
    \resizebox{0.79\columnwidth}{!}{%
        \begin{tabular}{cccccc}
        \toprule
        \# Prompt & Avg. AUC $\uparrow$       & Param. $\downarrow$        & Mem. $\downarrow$         & PPE $\uparrow$            & PME $\uparrow$            \\ \hline
        4         & 85.94          & \textbf{0.96} & \textbf{720} & 85.58          & 84.80          \\
        8         & 86.39          & \underline{0.99}    & \underline{722}    & 86.02          & 85.24          \\
        16        & \textbf{87.12} & 1.03          & 736          & \textbf{86.73} & \textbf{85.94} \\
        32        & 86.54          & 1.11          & 812          & 86.13          & \underline{85.25}    \\
        64        & 86.66          & 1.28          & 972         & \underline{86.18}    & 85.12          \\
        128       & 86.49          & 1.61          & 1,388         & 85.89          & 84.33          \\ \bottomrule
        \end{tabular}
    }
    \label{prompt}
\end{table}

\begin{table}[t]
    \vspace{-4pt}
    \centering
    \renewcommand{\arraystretch}{1.2}
    \caption{FPT+'s performance and efficiency with different reduction factors.}
    \resizebox{0.87\columnwidth}{!}{%
        \begin{tabular}{cccccc}
        \toprule
        Reduction Factor & Avg. AUC $\uparrow$          & Param. $\downarrow$          & Mem. $\downarrow$            & PPE $\uparrow$               & PME $\uparrow$               \\ \hline
        2                & 86.32             & 11.55           & 1,170            & 82.32             & 84.49             \\
        4                & \underline{86.74} & 3.40            & 894             & 85.49             & \underline{85.32} \\
        8                & \textbf{87.12}    & \underline{1.03} & \underline{736} & \textbf{86.73}    & \textbf{85.94}    \\
        16               & 85.92             & \textbf{0.37}   & \textbf{724}    & \underline{85.78} & 84.78             \\ \bottomrule
        \end{tabular}
    }
    \label{reduction_factor}
\end{table}

\begin{table}[!t]
    \vspace{-4pt}
    \centering
    \renewcommand{\arraystretch}{1.2}
    \caption{FPT+'s performance with different token selection strategies.}
    \resizebox{0.52\columnwidth}{!}{%
        \begin{tabular}{lc}
        \toprule
        Selection strategy & Avg. AUC $\uparrow$ \\
        \hline
        Random tokens selection & 85.79 \\
        Important tokens selection & \textbf{87.12} \\
        \bottomrule
        \end{tabular}
    }
    \label{selection}
\end{table}

\subsubsection{Reduction Factor}
The reduction factor $k$ in FPT+ determines the hidden dimension of the side network. As shown in Table~\ref{reduction_factor}, we assess the impact of different reduction factors on FPT+’s performance and efficiency. A reduction factor of 8 achieves the highest average AUC (87.12) with minimal memory usage (736 MB) and only 1.03\% of learnable parameters. Increasing the reduction factor to 16 slightly reduces memory usage but results in a drop in AUC (85.92), while lower factors (e.g., 2 and 4) increase resource consumption without significant gains in performance. We observe that a larger hidden dimension in the side network (lower reduction factor) does not improve performance as it increases the training complexity of the randomly initialized side network. A reduction factor of 8 achieves the optimal balance for high-resolution medical image classification tasks. 

\subsubsection{Token Selection Strategy}
To validate the effectiveness of the proposed important token selection strategy, we conduct an ablation study comparing it with a random token selection baseline. In the random strategy, tokens are selected uniformly at random rather than guided by attention-based importance scores. As presented in Table~\ref{selection}, important tokens selection consistently outperforms the random counterpart, achieving an average improvement of 1.33\% in AUC across all evaluated datasets. These results confirm that selecting diagnostically relevant tokens based on attention scores enables more efficient and effective transfer of pre-trained knowledge.

\begin{figure*}[t]
	\centering
	\includegraphics[width=\textwidth]{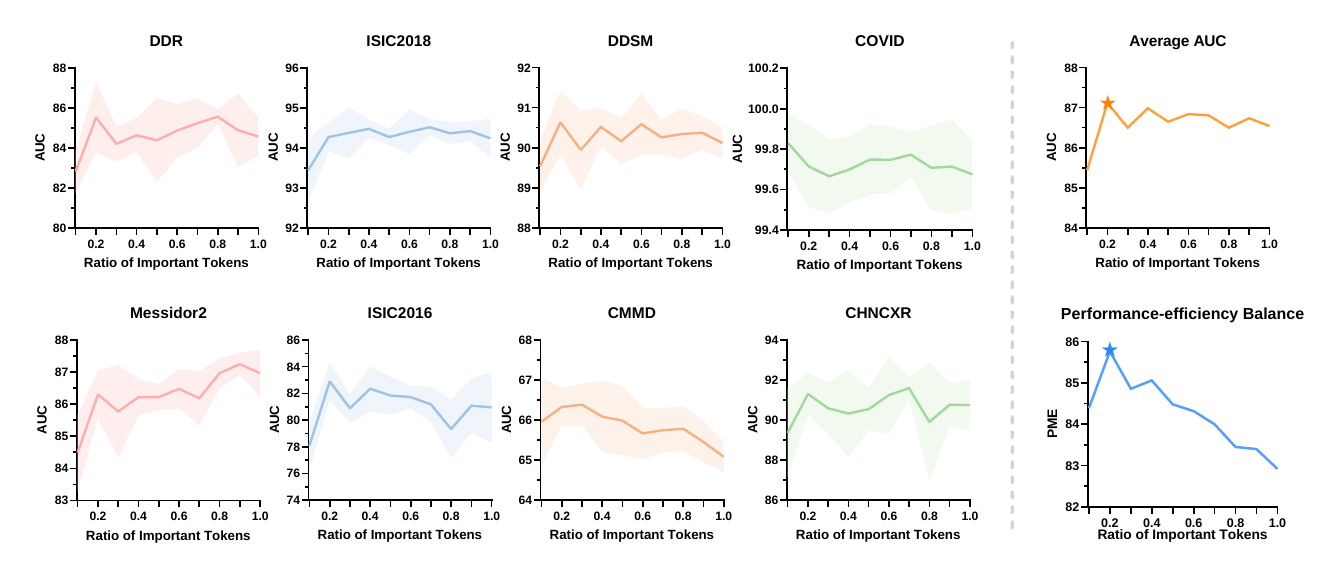}
	\caption{Impact of important tokens’ ratio on FPT+’s performance and efficiency with a 512 × 512 input resolution. The star symbol indicates the best one.}
	\label{ratio}
    \vspace{-4pt}
\end{figure*}

\begin{figure*}[t]
	\centering
	\includegraphics[width=\textwidth]{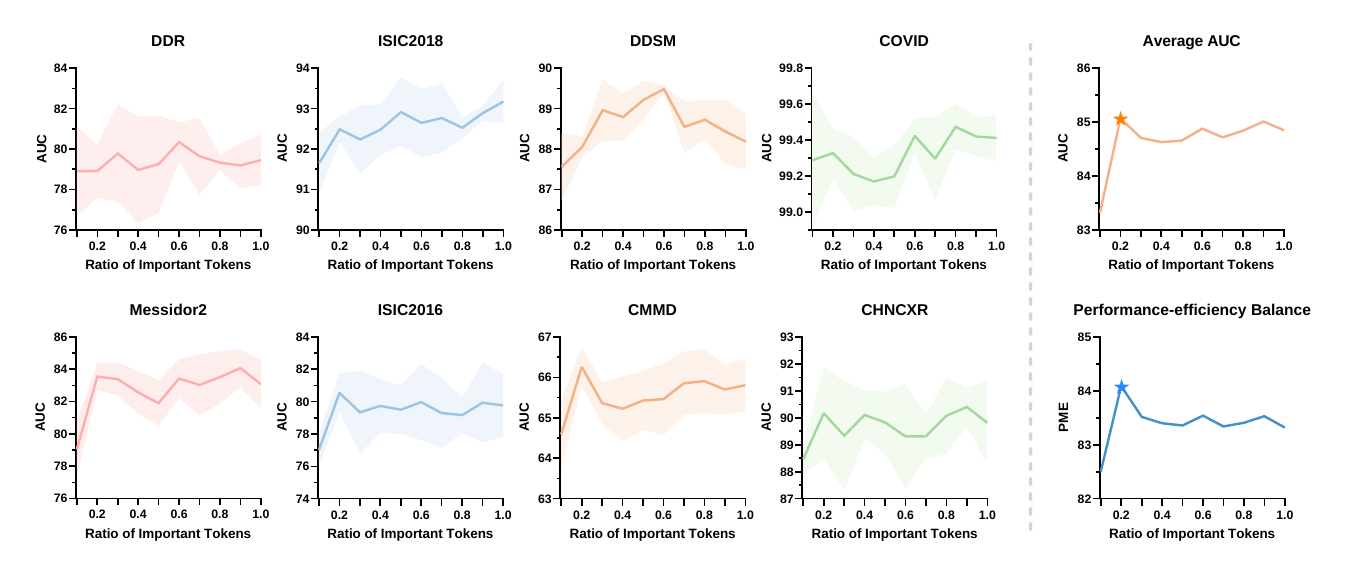}
	\caption{Impact of important tokens’ ratio on FPT+’s performance and efficiency with a 256 × 256 input resolution. The star symbol indicates the best one.}
	\label{ratio}
    \vspace{-4pt}
\end{figure*}

\subsubsection{Ratio of Important Tokens}
To mitigate the training cost, we exclusively integrate features derived from important tokens of the pre-trained model into the side network. In this section, we assess how different ratios of these important tokens affect the overall performance and efficiency. We provide a detailed analysis of FPT+ under varying token ratios, focusing on both 512 × 512 and 256 × 256 input resolutions. We notice that the optimal ratio of important tokens varies across different datasets. For instance, although both DDR and Messidor-2 are fundus image datasets, they show peak performance at different important token ratios. Although a higher ratio of important tokens can provide more information for fusion modules, it carries a risk of misleading the fusion modules and potentially increasing the difficulty of learning. Most datasets prefer a token ratio lower than 0.5, except for Messidor-2 and CHNCXR which achieve the best performance at 0.9 and 0.7, respectively. We conjecture that the optimal token ratio depends on the scale and complexity of the dataset of interest.

In terms of memory efficiency, as the ratio of important tokens increases, the memory consumption also rises dramatically, resulting in significant drops in PME. Across both 512 × 512 and 256 × 256 input resolutions, a token ratio of 0.2 consistently achieves the best trade-off, yielding the highest average AUC while maintaining low memory usage. Although FPT+'s performance may benefit from dataset-specific tuning on important token ratio, a 0.2 token ratio offers a stable and effective choice in a wide range of scenarios.

\subsection{Visualization}
\begin{figure*}[t]
	\centering
	\includegraphics[width=0.98\textwidth]{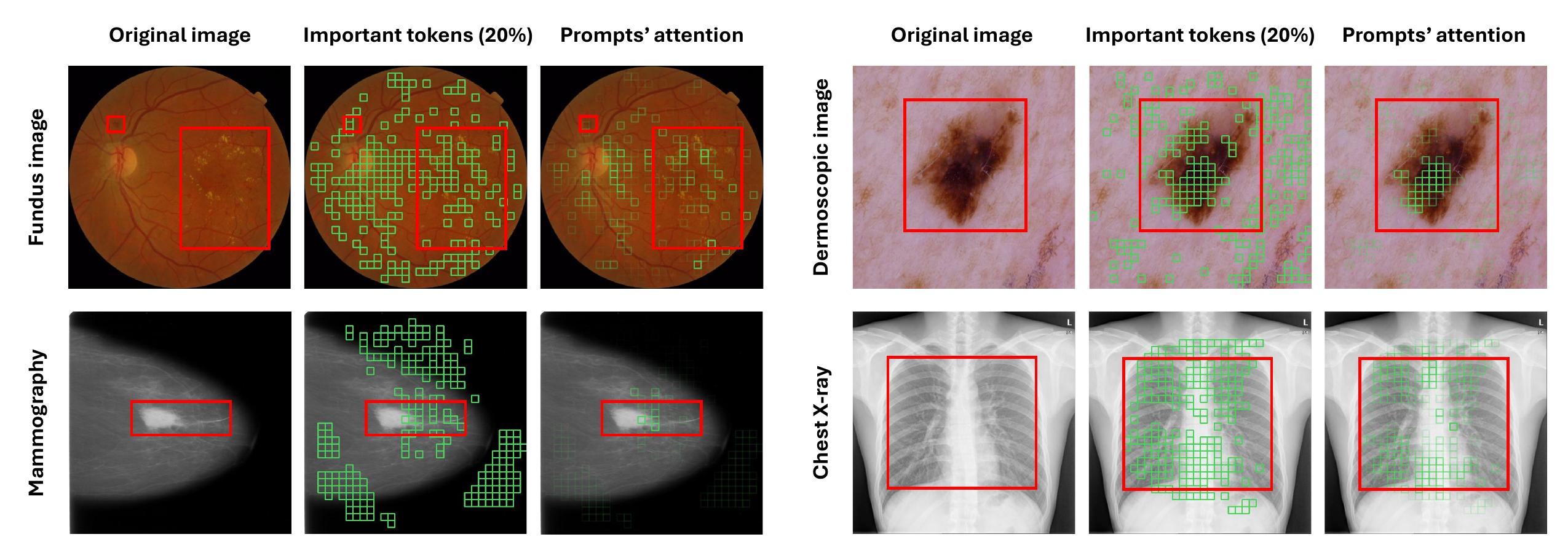}
	\caption{Visualization of the selected important tokens and attention scores from the fine-grained prompts for representative cases. Clinically relevant areas are highlighted by red bounding boxes.}
	\label{visualization}
\end{figure*}

To verify the important tokens selection and the mechanism of the fusion modules, we visualize the selected tokens of the top layer and the attention scores on them in the corresponding fusion module across four modalities. As shown in Fig.~\ref{visualization}, the selected important tokens are highlighted in green boxes and clinically relevant areas are highlighted by red bounding boxes. It can be observed that the selected tokens cover most discriminative regions, such as lesions and the optic disc in fundus images. Please note that ViT employs a multi-head mechanism, and the visualization is only for one of the multiple heads. This multi-head mechanism ensures that the important tokens selection can convey sufficient diagnostic clues to the side network. Additionally, we visualize the attention that the fine-grained prompts pay to the selected tokens. It can be observed that the fine-grained prompts further refine the selection, reducing redundancy and summarizing them into a small set of prompts for the side network. This refinement process ensures that only the most relevant information is conveyed, enhancing the efficiency and effectiveness of the transfer learning process.

\section{Discussion}
With the success of large-scale pre-trained models, transfer learning has become a key technique to apply pre-trained knowledge in real-world applications. PETL has been proposed to efficiently adapt LPMs to new downstream tasks. However, existing PETL methods mainly focus on reducing the number of learnable parameters, and the improvements in memory efficiency are not as effective as those in parameter efficiency. Medical image analysis is a domain that often requires high-resolution inputs to ensure that tiny diagnostic clues are clearly exposed. However, increasing the input resolution dramatically increases memory consumption during training. As shown in Fig.~\ref{memory}, even the most memory-efficient PETL method, linear probing, is not feasible when the input resolution is $1280 \times 1280$ on a GPU with a 48GB memory size. Our proposed FPT+ significantly improves memory efficiency, making training at $2,048 \times 2,048$ possible on a single GPU, with the potential for even higher resolutions based on the trend shown in the figure.

Although FPT+ makes the training of super high-resolution images feasible on one GPU, we notice that the performance saturates or even slightly drops when the input resolution exceeds $1,024 \times 1,024$. There may be two plausible reasons: (1) The LPM used is pre-trained at a standard resolution of $224 \times 224$, and its feature extraction ability degrades when the resolution is significantly different from the pre-trained resolution; (2) The datasets we use may have already been resized and compressed by their owners, allowing users to download them from the website. Hence, increasing the input size to go beyond the original image size shall not provide any additional information.

High-resolution image classification is not the only field that benefits from this memory efficiency. The memory efficiency of FPT+ allows for larger batch sizes during training, which can accelerate the training process and potentially improve the stability and performance of transfer learning. Additionally, there are settings in natural image analysis that also require high resolution, such as satellite imagery \cite{Adegun2023}. The generalizability of FPT+ to natural images remains unexplored, which is one of our future works.

\section{Conclusion}
In this paper, we extend our previous work FPT to FPT+. FPT+ significantly improves the memory efficiency over existing PETL methods, particularly in the high-resolution context commonly encountered in medical image analysis. To reduce memory consumption, we first enhance the design of side tuning with asymmetric input. To effectively adapt pre-trained knowledge from the LPM, we introduce fine-grained prompts and fusion modules to summarize and process intermediate features of the LPM and effectively convey them to the side network. To further reduce memory utilization, we propose the selection of important tokens and the preloading of pre-trained features. By integrating all these components, we present a parameter and memory-efficient transfer learning method for high-resolution medical image classification. The effectiveness of FPT+ is successfully established across eight medical image datasets of diverse sizes, modalities, and complexities. Experimental results show that FPT+ outperforms all other SOTA PETL methods while maintaining the best parameter and memory efficiency.

\balance
\bibliographystyle{plain}
\bibliography{./ref.bib}

\begin{thebibliography}{10}

\bibitem{Adegun2023}
Adekanmi~Adeyinka Adegun, Serestina Viriri, and Jules-Raymond Tapamo.
\newblock Review of deep learning methods for remote sensing satellite images classification: experimental survey and comparative analysis.
\newblock {\em Journal of Big Data}, 10:93, 2023.

\bibitem{Elad2022}
Elad Ben-Zaken, Shauli Ravfogel, and Yoav Goldberg.
\newblock Bitfit: Simple parameter-efficient fine-tuning for transformer-based masked language-models.
\newblock In {\em Proceedings of the Annual Meeting of the Association for Computational Linguistics}, volume~2, 2022.

\bibitem{brown2020language}
Tom Brown, Benjamin Mann, Nick Ryder, Melanie Subbiah, Jared~D Kaplan, Prafulla Dhariwal, Arvind Neelakantan, Pranav Shyam, Girish Sastry, Amanda Askell, et~al.
\newblock Language models are few-shot learners.
\newblock {\em Advances in neural information processing systems}, 33:1877--1901, 2020.

\bibitem{10580962}
Zhiyuan Cai, Li~Lin, Huaqing He, Pujin Cheng, and Xiaoying Tang.
\newblock Uni4eye++: A general masked image modeling multi-modal pre-training framework for ophthalmic image classification and segmentation.
\newblock {\em IEEE Transactions on Medical Imaging}, pages 1--1, 2024.

\bibitem{Caron2021}
Mathilde Caron, Hugo Touvron, Ishan Misra, Hervé Jégou, Julien Mairal, Piotr Bojanowski, and Armand Joulin.
\newblock Emerging properties in self-supervised vision transformers.
\newblock In {\em Proceedings of the IEEE/CVF international conference on computer vision}, pages 9650--9660, 2021.

\bibitem{Chen2021}
Chun-Fu~Richard Chen, Quanfu Fan, and Rameswar Panda.
\newblock Crossvit: Cross-attention multi-scale vision transformer for image classification.
\newblock In {\em Proceedings of the IEEE/CVF international conference on computer vision}, pages 357--366, 2021.

\bibitem{Chen2022}
Zhe Chen, Yuchen Duan, Wenhai Wang, Junjun He, Tong Lu, Jifeng Dai, and Yu~Qiao.
\newblock Vision transformer adapter for dense predictions.
\newblock {\em ICLR 2023}, 2022.

\bibitem{Cheng2023}
Pujin Cheng, Li~Lin, Junyan Lyu, Yijin Huang, Wenhan Luo, and Xiaoying Tang.
\newblock Prior: Prototype representation joint learning from medical images and reports.
\newblock In {\em Proceedings of the IEEE/CVF International Conference on Computer Vision}, pages 21361--21371, 2023.

\bibitem{Codella2019}
Noel Codella, Veronica Rotemberg, Philipp Tschandl, M~Emre Celebi, Stephen Dusza, David Gutman, Brian Helba, Aadi Kalloo, Konstantinos Liopyris, Michael Marchetti, et~al.
\newblock Skin lesion analysis toward melanoma detection 2018: A challenge hosted by the international skin imaging collaboration (isic).
\newblock {\em arXiv preprint arXiv:1902.03368}, 2019.

\bibitem{Cui2021}
Chunyan Cui, Li~Li, Hongmin Cai, Zhihao Fan, Ling Zhang, Tingting Dan, Jiao Li, and Jinghua Wang.
\newblock The chinese mammography database (cmmd): An online mammography database with biopsy confirmed types for machine diag- nosis of breast.
\newblock {\em Data from The Cancer Imaging Archive.}, 2021.

\bibitem{Messidor2}
Etienne Decencière, Xiwei Zhang, Guy Cazuguel, Bruno Lay, Béatrice Cochener, Caroline Trone, Philippe Gain, Richard Ordonez, Pascale Massin, Ali Erginay, et~al.
\newblock Feedback on a publicly distributed image database: the messidor database.
\newblock {\em Image Analysis and Stereology}, 33:231--234, 2014.

\bibitem{Deng2009}
Jia Deng, Wei Dong, Richard Socher, Li-Jia Li, Kai Li, and Li~Fei-Fei.
\newblock Imagenet: A large-scale hierarchical image database.
\newblock In {\em 2009 IEEE conference on computer vision and pattern recognition}, pages 248--255, 2009.

\bibitem{Devlin2019}
Jacob Devlin, Ming~Wei Chang, Kenton Lee, and Kristina Toutanova.
\newblock Bert: Pre-training of deep bidirectional transformers for language understanding.
\newblock In {\em NAACL HLT 2019 - 2019 Conference of the North American Chapter of the Association for Computational Linguistics: Human Language Technologies - Proceedings of the Conference}, volume~1, 2019.

\bibitem{Ding2023}
Ning Ding, Yujia Qin, Guang Yang, Fuchao Wei, Zonghan Yang, Yusheng Su, Shengding Hu, Yulin Chen, Chi~Min Chan, Weize Chen, Jing Yi, Weilin Zhao, Xiaozhi Wang, Zhiyuan Liu, Hai~Tao Zheng, Jianfei Chen, Yang Liu, Jie Tang, Juanzi Li, and Maosong Sun.
\newblock Parameter-efficient fine-tuning of large-scale pre-trained language models.
\newblock {\em Nature Machine Intelligence}, 5, 2023.

\bibitem{Dosovitskiy2021}
Alexey Dosovitskiy, Lucas Beyer, Alexander Kolesnikov, Dirk Weissenborn, Xiaohua Zhai, Thomas Unterthiner, Mostafa Dehghani, Matthias Minderer, Georg Heigold, Sylvain Gelly, Jakob Uszkoreit, and Neil Houlsby.
\newblock An image is worth 16x16 words: Transformers for image recognition at scale.
\newblock In {\em ICLR 2021 - 9th International Conference on Learning Representations}, 2021.

\bibitem{Dutt2023}
Raman Dutt, Linus Ericsson, Pedro Sanchez, Sotirios~A Tsaftaris, and Timothy Hospedales.
\newblock Parameter-efficient fine-tuning for medical image analysis: The missed opportunity.
\newblock {\em arXiv preprint arXiv:2305.08252}, 2023.

\bibitem{Gong2023}
Shizhan Gong, Yuan Zhong, Wenao Ma, Jinpeng Li, Zhao Wang, Jingyang Zhang, Pheng-Ann Heng, and Qi~Dou.
\newblock 3dsam-adapter: Holistic adaptation of sam from 2d to 3d for promptable medical image segmentation.
\newblock {\em arXiv preprint arXiv:2306.13465}, 2023.

\bibitem{Gutman2016}
David Gutman, Noel C~F Codella, Emre Celebi, Brian Helba, Michael Marchetti, Nabin Mishra, and Allan Halpern.
\newblock Skin lesion analysis toward melanoma detection: A challenge at the international symposium on biomedical imaging (isbi) 2016, hosted by the international skin imaging collaboration (isic).
\newblock {\em arXiv preprint arXiv:1605.01397}, 2016.

\bibitem{He2023DVPT}
Along He, Kai Wang, Zhihong Wang, Tao Li, and Huazhu Fu.
\newblock Dvpt: Dynamic visual prompt tuning of large pre-trained models for medical image analysis.
\newblock {\em arXiv preprint arXiv:2307.09787}, 7 2023.

\bibitem{He2023}
Xuehai He, Chuanyuan Li, Pengchuan Zhang, Jianwei Yang, and Xin~Eric Wang.
\newblock Parameter-efficient model adaptation for vision transformers.
\newblock In {\em Proceedings of the 37th AAAI Conference on Artificial Intelligence, AAAI 2023}, volume~37, 2023.

\bibitem{Houlsby2019}
Neil Houlsby, Andrei Giurgiu, Stanislaw Jastrzebski, Bruna Morrone, Quentin~De Laroussilhe, Andrea Gesmundo, Mona Attariyan, and Sylvain Gelly.
\newblock Parameter-efficient transfer learning for nlp.
\newblock In {\em International Conference on Machine Learning}, pages 2790--2799, 2019.

\bibitem{Hu2022}
Edward Hu, Yelong Shen, Phillip Wallis, Zeyuan Allen-Zhu, Yuanzhi Li, Shean Wang, Lu~Wang, and Weizhu Chen.
\newblock Lora: Low-rank adaptation of large language models.
\newblock In {\em ICLR 2022 - 10th International Conference on Learning Representations}, 2022.

\bibitem{Huang2024FPT}
Yijin Huang, Pujin Cheng, Roger Tam, and Xiaoying Tang.
\newblock Fine-grained prompt tuning: A parameter and memory efficient transfer learning method for high-resolution medical image classification.
\newblock {\em arXiv preprint arXiv:2403.07576}, 3 2024.

\bibitem{Huang2023}
Yijin Huang, Li~Lin, Pujin Cheng, Junyan Lyu, Roger Tam, and Xiaoying Tang.
\newblock Identifying the key components in resnet-50 for diabetic retinopathy grading from fundus images: a systematic investigation.
\newblock {\em Diagnostics}, 13:1664, 2023.

\bibitem{Huang2024}
Yijin Huang, Junyan Lyu, Pujin Cheng, Roger Tam, and Xiaoying Tang.
\newblock Ssit: Saliency-guided self-supervised image transformer for diabetic retinopathy grading.
\newblock {\em IEEE Journal of Biomedical and Health Informatics}, 2024.

\bibitem{Jaeger2014}
Stefan Jaeger, Sema Candemir, Sameer Antani, Y\`\i-Xiáng~J Wáng, Pu-Xuan Lu, and George Thoma.
\newblock Two public chest x-ray datasets for computer-aided screening of pulmonary diseases.
\newblock {\em Quantitative imaging in medicine and surgery}, 4:475, 2014.

\bibitem{Jia2022}
Menglin Jia, Luming Tang, Bor-Chun Chen, Claire Cardie, Serge Belongie, Bharath Hariharan, and Ser-Nam Lim.
\newblock Visual prompt tuning.
\newblock In {\em European Conference on Computer Vision}, pages 709--727, 2022.

\bibitem{Kim2024}
Yumin Kim, Gayoon Choi, and Seong~Jae Hwang.
\newblock Parameter efficient fine tuning for multi-scanner pet to pet reconstruction, 2024.

\bibitem{Kirillov2023}
Alexander Kirillov, Eric Mintun, Nikhila Ravi, Hanzi Mao, Chloe Rolland, Laura Gustafson, Tete Xiao, Spencer Whitehead, Alexander~C Berg, Wan-Yen Lo, Piotr Dollár, and Ross Girshick.
\newblock Segment anything.
\newblock In {\em 2023 IEEE/CVF International Conference on Computer Vision (ICCV)}, pages 3992--4003. IEEE, 7 2023.

\bibitem{Lee2017}
Rebecca~Sawyer Lee, Francisco Gimenez, Assaf Hoogi, Kanae~Kawai Miyake, Mia Gorovoy, and Daniel~L Rubin.
\newblock A curated mammography data set for use in computer-aided detection and diagnosis research.
\newblock {\em Scientific data}, 4:1--9, 2017.

\bibitem{Lester2021}
Brian Lester, Rami Al-Rfou, and Noah Constant.
\newblock The power of scale for parameter-efficient prompt tuning.
\newblock In {\em EMNLP 2021 - 2021 Conference on Empirical Methods in Natural Language Processing, Proceedings}, 2021.

\bibitem{Li2022}
Chunyuan Li, Haotian Liu, Liunian Li, Pengchuan Zhang, Jyoti Aneja, Jianwei Yang, Ping Jin, Houdong Hu, Zicheng Liu, Yong~Jae Lee, et~al.
\newblock Elevater: A benchmark and toolkit for evaluating language-augmented visual models.
\newblock {\em Advances in Neural Information Processing Systems}, 35:9287--9301, 2022.

\bibitem{Li2019}
Tao Li, Yingqi Gao, Kai Wang, Song Guo, Hanruo Liu, and Hong Kang.
\newblock Diagnostic assessment of deep learning algorithms for diabetic retinopathy screening.
\newblock {\em Information Sciences}, 501:511--522, 2019.

\bibitem{Liu2023}
Weihuang Liu, Xi~Shen, Chi~Man Pun, and Xiaodong Cun.
\newblock Explicit visual prompting for low-level structure segmentations.
\newblock In {\em Proceedings of the IEEE Computer Society Conference on Computer Vision and Pattern Recognition}, volume 2023-June, 2023.

\bibitem{Ma2024}
Jun Ma, Yuting He, Feifei Li, Lin Han, Chenyu You, and Bo~Wang.
\newblock Segment anything in medical images.
\newblock {\em Nature Communications}, 15, 2024.

\bibitem{paszke2017automatic}
Adam Paszke, Sam Gross, Soumith Chintala, Gregory Chanan, Edward Yang, Zachary DeVito, Zeming Lin, Alban Desmaison, Luca Antiga, and Adam Lerer.
\newblock Automatic differentiation in pytorch.
\newblock 2017.

\bibitem{Radford2021}
Alec Radford, Jong~Wook Kim, Chris Hallacy, Aditya Ramesh, Gabriel Goh, Sandhini Agarwal, Girish Sastry, Amanda Askell, Pamela Mishkin, Jack Clark, et~al.
\newblock Learning transferable visual models from natural language supervision.
\newblock In {\em International conference on machine learning}, pages 8748--8763, 2021.

\bibitem{Radford2019}
Alec Radford, Jeffrey Wu, Rewon Child, David Luan, Dario Amodei, Ilya Sutskever, et~al.
\newblock Language models are unsupervised multitask learners.
\newblock {\em OpenAI blog}, 1:9, 2019.

\bibitem{raghu2021vision}
Maithra Raghu, Thomas Unterthiner, Simon Kornblith, Chiyuan Zhang, and Alexey Dosovitskiy.
\newblock Do vision transformers see like convolutional neural networks?
\newblock {\em Advances in neural information processing systems}, 34:12116--12128, 2021.

\bibitem{Ridnik2021}
Tal Ridnik, Emanuel Ben-Baruch, Asaf Noy, and Lihi Zelnik-Manor.
\newblock Imagenet-21k pretraining for the masses.
\newblock {\em arXiv preprint arXiv:2104.10972}, 2021.

\bibitem{Shen2017}
Dinggang Shen, Guorong Wu, and Heung-Il Suk.
\newblock Deep learning in medical image analysis.
\newblock {\em Annual review of biomedical engineering}, 19:221--248, 2017.

\bibitem{Siddhartha2021}
Manu Siddhartha.
\newblock Covid cxr image dataset (research), 2021.

\bibitem{Julio2023}
Julio Silva-Rodriguez, Jose Dolz, and Ismail~Ben Ayed.
\newblock Transductive few-shot adapters for medical image segmentation.
\newblock {\em arXiv preprint arXiv:2303.17051}, 2023.

\bibitem{Sung2022}
Yi-Lin Sung, Jaemin Cho, and Mohit Bansal.
\newblock Lst: Ladder side-tuning for parameter and memory efficient transfer learning.
\newblock {\em Advances in Neural Information Processing Systems}, 35:12991--13005, 2022.

\bibitem{Touvron2022}
Hugo Touvron, Matthieu Cord, Alaaeldin El-Nouby, Jakob Verbeek, and Hervé Jégou.
\newblock Three things everyone should know about vision transformers.
\newblock In {\em European Conference on Computer Vision}, pages 497--515, 2022.

\bibitem{Vaswani2017}
Ashish Vaswani, Noam Shazeer, Niki Parmar, Jakob Uszkoreit, Llion Jones, Aidan~N Gomez, Łukasz Kaiser, and Illia Polosukhin.
\newblock Attention is all you need.
\newblock {\em Advances in neural information processing systems}, 30, 2017.

\bibitem{Weiss2016}
Karl Weiss, Taghi~M Khoshgoftaar, and DingDing Wang.
\newblock A survey of transfer learning.
\newblock {\em Journal of Big data}, 3:1--40, 2016.

\bibitem{Wu2023}
Junde Wu, Wei Ji, Yuanpei Liu, Huazhu Fu, Min Xu, Yanwu Xu, and Yueming Jin.
\newblock Medical sam adapter: Adapting segment anything model for medical image segmentation.
\newblock {\em arXiv preprint arXiv:2304.12620}, 4 2023.

\bibitem{Yang2019}
Zhilin Yang, Zihang Dai, Yiming Yang, Jaime Carbonell, Russ~R Salakhutdinov, and Quoc~V Le.
\newblock Xlnet: Generalized autoregressive pretraining for language understanding.
\newblock {\em Advances in neural information processing systems}, 32, 2019.

\bibitem{Zhang2020}
Jeffrey~O Zhang, Alexander Sax, Amir Zamir, Leonidas Guibas, and Jitendra Malik.
\newblock Side-tuning: a baseline for network adaptation via additive side networks.
\newblock In {\em Computer Vision–ECCV 2020: 16th European Conference, Glasgow, UK, August 23–28, 2020, Proceedings, Part III 16}, pages 698--714, 2020.

\bibitem{Zhang2023}
Qingru Zhang, Minshuo Chen, Alexander Bukharin, Nikos Karampatziakis, Pengcheng He, Yu~Cheng, Weizhu Chen, and Tuo Zhao.
\newblock Adalora: Adaptive budget allocation for parameter-efficient fine-tuning.
\newblock {\em arXiv preprint arXiv:2303.10512}, 3 2023.

\bibitem{Zhang2023biomedclip}
Sheng Zhang, Yanbo Xu, Naoto Usuyama, Hanwen Xu, Jaspreet Bagga, Robert Tinn, Sam Preston, Rajesh Rao, Mu~Wei, Naveen Valluri, Cliff Wong, Andrea Tupini, Yu~Wang, Matt Mazzola, Swadheen Shukla, Lars Liden, Jianfeng Gao, Matthew~P. Lungren, Tristan Naumann, Sheng Wang, and Hoifung Poon.
\newblock Biomedclip: a multimodal biomedical foundation model pretrained from fifteen million scientific image-text pairs.
\newblock {\em arXiv preprint arXiv:2303.00915}, 3 2023.

\bibitem{Zhao2024}
Henry~Hengyuan Zhao, Pichao Wang, Yuyang Zhao, Hao Luo, Fan Wang, and Mike~Zheng Shou.
\newblock Sct: A simple baseline for parameter-efficient fine-tuning via salient channels.
\newblock {\em International Journal of Computer Vision}, 132, 2024.

\bibitem{Zhou2022cocoop}
Kaiyang Zhou, Jingkang Yang, Chen~Change Loy, and Ziwei Liu.
\newblock Conditional prompt learning for vision-language models.
\newblock In {\em Proceedings of the IEEE Computer Society Conference on Computer Vision and Pattern Recognition}, volume 2022-June, 2022.

\bibitem{Zhou2022}
Kaiyang Zhou, Jingkang Yang, Chen~Change Loy, and Ziwei Liu.
\newblock Learning to prompt for vision-language models.
\newblock {\em International Journal of Computer Vision}, 130, 2022.

\bibitem{zhou2023foundation}
Yukun Zhou, Mark~A Chia, Siegfried~K Wagner, Murat~S Ayhan, Dominic~J Williamson, Robbert~R Struyven, Timing Liu, Moucheng Xu, Mateo~G Lozano, Peter Woodward-Court, et~al.
\newblock A foundation model for generalizable disease detection from retinal images.
\newblock {\em Nature}, 622(7981):156--163, 2023.

\bibitem{Zu2024}
Wenqiang Zu, Shenghao Xie, Qing Zhao, Guoqi Li, and Lei Ma.
\newblock Embedded prompt tuning: Towards enhanced calibration of pretrained models for medical images.
\newblock {\em Medical Image Analysis}, page 103258, 2024.

\end{thebibliography}

\vspace{-30pt}
\begin{IEEEbiography}[{\includegraphics[width=1in,height=1.25in,clip,keepaspectratio]{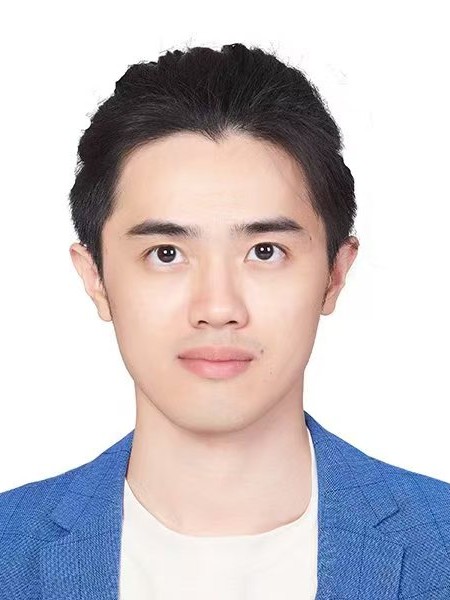}}]{Yijin Huang} received the B.S. degrees from the Southern University of Science and Technology, Shenzhen, China; M.S. degrees from Carnegie Mellon University, Pittsburgh, USA. He is currently pursuing the Ph.D. degree with School of Biomedical Engineering, The University of British Columbia, Vancouver, Canada.

His current research interests include Self-supervised pre-training, vision-language model pre-training, and parameter-efficient fine-tuning.\end{IEEEbiography}

\vspace{-30pt}
\begin{IEEEbiography}[{\includegraphics[width=1in,height=1.25in,clip,keepaspectratio]{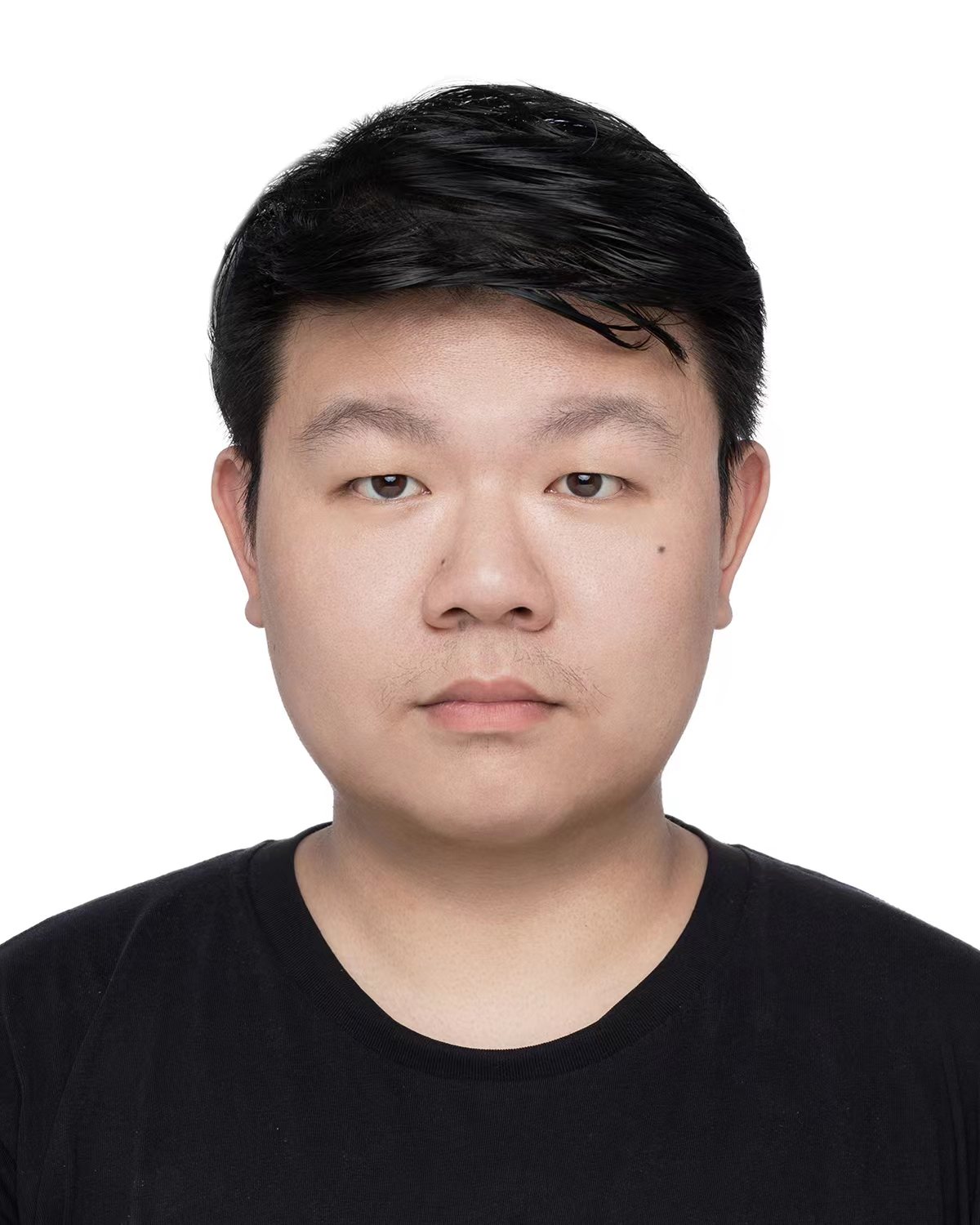}}]{Pujin Cheng} received the B.S. and M.S. degrees from the Southern University of Science and Technology, Shenzhen, China. He is currently pursuing the Ph.D. degree with the Department of Electrical and Electronic Engineering, the University of Hong Kong, Hong Kong, China.

His current research interests include vision-language model pre-training, interpretable learning, and medical data mining.\end{IEEEbiography}

\vspace{-30pt}
\begin{IEEEbiography}[{\includegraphics[width=1in,height=1.25in,clip,keepaspectratio]{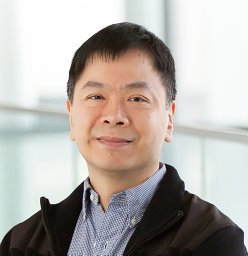}}]{Roger Tam} received the Ph.D. degree in computer science from the University of British Columbia (UBC), Vancouver, Canada.

Dr. Tam is currently an Associate Professor in the Department of Radiology at UBC and a member of the MS/MRI Research Group in the Division of Neurology.  His research interests are centered on the application of computer vision and machine learning methods to the quantitative analysis of medical images. \end{IEEEbiography}

\vspace{-30pt}
\begin{IEEEbiography}[{\includegraphics[width=1in,height=1.25in,clip,keepaspectratio]{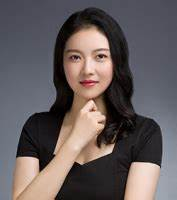}}]{Xiaoying Tang} received the Ph.D. degree in electrical and computer engineering from Johns Hopkins University, Baltimore, USA.

Dr. Tang is currently an Associate Professor and Researcher in the Department of Electronic and Electrical Engineering at the Southern University of Science and Technology. Her research focuses on intelligent medical image computation and analysis, with primary applications to the multi-modality MRI image analysis of the human brain and the multi-modality ophthalmic images. \end{IEEEbiography}

\end{document}